# Supporting Migration Policies with Forecasts: Illegal Border Crossings in Europe through a Mixed Approach

**Authors**

Claudio Bosco*[1], Umberto Minora[1], Daniele de Rigo[2], Julia Pingsdorf[1], Roberto Cortinovis[1]

[1]European Commission - Joint Research Centre, Via Enrico Fermi 2749, Ispra, VA, Italy

[2]Maieutike Research Initiative, Milan, MI, Italy.

*corresponding author: claudio.bosco@ec.europa.eu

orcid Claudio Bosco: https://orcid.org/0000-0002-6438-4571

orcid Umberto Minora: https://orcid.org/0000-0003-2151-6500

orcid Daniele De Rigo: https://orcid.org/0000-0003-0863-2670

## Abstract

This paper presents a mixed-methodology to forecast illegal border crossings in Europe across five key migratory routes, with a one-year time horizon. The methodology integrates machine learning techniques with qualitative insights from migration experts. This approach aims at improving the predictive capacity of data-driven models through the inclusion of a human-assessed covariate, an innovation that addresses challenges posed by sudden shifts in migration patterns and limitations in traditional datasets.

The proposed methodology responds directly to the forecasting needs outlined in the EU Pact on Migration and Asylum, supporting the Asylum and Migration Management Regulation (AMMR). It is designed to provide policy-relevant forecasts that inform strategic decisions, early warning systems, and solidarity mechanisms among EU Member States.

By joining data-driven modeling with expert judgment, this work aligns with existing academic recommendations and introduces a novel operational tool tailored for EU migration governance. The methodology is tested and validated with known data to demonstrate its applicability and reliability in migration-related policy context.

# 1 Introduction

In the present work, we use a mixed methodology to forecast one-year ahead illegal border-crossings to the European Union through five migratory routes (as defined by Frontex, https://www.frontex.europa.eu/what-we-do/monitoring-and-risk-analysis/migratory-routes/migratory-routes). This "mixed" methodology combines quantitative data with qualitative inputs from experienced analysts about potential future developments. The predictions are in the form of aggregated prediction ranges, thus providing a unique minimum and maximum value for the whole forecasting time horizon by route.

This work represents a revision and improvement of the methodology presented in Bosco et al. [1]. Building on top of the previous methodology, the authors makes use of machine learning to generate predictions, but instead of relying on a purely data-driven solution they included a human assessed covariate in the model to improve its outcome (more on this in Section 2.1). The inclusion of such variable helps the model to overcome the challenges and limitations caused by the lack of signal in the traditional migration datasets about sudden changing migratory regimes.

The inclusion of expert knowledge in migration forecasts was already suggested in Willekens [2]. Many researchers have already attempted to forecast migration using a variety of techniques and focusing on different scenarios, time horizons, regions, and aspects of migration. Boissonneault et al. [3] provide an overview on this topic, based on the analysis of 107 documents. The authors divided the focus of the selected documents between either migration or other aspects of societies influenced by migration. In their review, they state that only one-third of the selected documents focus on migration as outcome, and, of these, just four use a mixed approach (quantitative and qualitative) at building migration scenario. According to Table A3 in the same literature review, Bijak & Wisniowski [4] is the only work that, similar to the current research, relies both on time series analysis (past trend) and on an own participatory work, where specialists and stakeholders are brought together to elaborate narratives of migration futures. Their main conclusions are that i) migration is barely predictable, mainly for the non-stationary nature of the data and the lack of data or long time series; ii) uncertainty matters, as a consequence of the previous point and because this would give a more realistic view if informed decisions are to be made; iii) expert knowledge matters, although the effect of subjective expertise was much less profound with respect to the determination of the nature of the processes under study; iv) forecasts with too long horizons are useless, due to the often non-stationary nature of the processes under study and the consequent increasing uncertainty over time. Here we propose a methodology that shares the "mixed" nature with Bijak & Wisniowski [4] but is profoundly different in the model choice and type of qualitative inputs, and tries to tackle these main issues with an innovative approach.

The methodology presented in this paper was developed with policy support in mind, as it aims to address the forecasting requirements of the Pact on Migration and Asylum, adopted on 14 May 2024 (https://home-affairs.ec.europa.eu/policies/migration-and-asylum/pact-migration-and-asylum_en). The Pact establishes a new framework for migration management at the EU level and for governing the Common European Asylum System. One of its key instruments, the Asylum and Migration Management Regulation (AMMR)[5], mandates the European Commission to adopt a European Annual Asylum and Migration Management Report by the 15th October each year. This comprehensive Report assesses the asylum, reception and migratory situation over the preceding 12 months along all migratory routes and Member States. As such, it serves early warning and awareness purposes, and provides a strategic situational picture which should include forecasting for the following year. Specifically, Article 9(b) of the AMMR stipulates that the Report should include "a projection for the coming year, including the number of anticipated arrivals by sea, based on the overall migratory situation in the previous year and considering the current situation, while also reflecting previous pressure".

The Report provides a critical evidence base for migration management at the EU level, serving as a key resource for supporting policy decisions. It informs the Commission's Implementing Decision to assess whether a Member State is experiencing migratory pressure, risk of migratory pressure or a significant migratory situation; and underpins the Commission's proposal for a Council implementing act establishing the Annual Solidarity Pool. The Pact introduces a permanent mandatory solidarity mechanism allowing Member States under migratory pressure to access support in form of relocations, financial contributions, or alternative solidarity measures. Member States facing significant migratory situations can in turn benefit from partial or full deductions to their solidarity contributions.

The purpose of this paper is to thoroughly explain the methodology we developed for forecasting migration patterns and to validate this approach using established, known data, proving its effectiveness and reliability in real-world applications.

## 2 Methods

### 2.1 Data

#### 2.1.1 Illegal Border Crossings

Illegal border crossings (IBCs) are the dependent variable considered in our study. These data are publicly available on the web. The forecasting model described here has been tested with publicly available monthly data from Frontex (the European Border and Coast Guard Agency) website (https://www.frontex.europa.eu/what-we-do/monitoring-and-risk-analysis/migratory-map). These are updated on a monthly basis and are available since January 2009. They are disaggregated by different Migratory Routes as defined by Frontex. Here we focused on the total number of IBCs on five migratory routes: Central Mediterranean route (CMR), Eastern Mediterranean route (EMR), Western Mediterranean route (WMR), West African (or Atlantic) route (WAR) and Western Balkans route (WBR), without distinguishing between different types of borders or the nationality of intercepted individuals.

Data on IBCs describe detections of illegal border-crossings on entry between border crossing points (BCPs) of the EU's external borders. It is important to underline that data on IBCs do not equate to the number of persons detected, the same person may cross the external border several times or she/he may cross the border without being detected.

No missing values are present within the datasets used for this exercise. However, it is important to mention that the latest data available in the datasets is generally 2 to 3 months behind the current date of forecasting. If data has a publication delay of (e.g.) 3 months, making a 12-months ahead forecasting would mean to extend the forecasting horizon to 15 months.

## 2.1.2 Defining a suite of covariates for predicting IBCs

Choosing the optimal set of covariates is crucial for maximizing a model's forecasting capacity. Including too few informative covariates could limit the model's explanatory power, while incorporating too many covariates can lead to overfitting, which means that the model is not able to 'learn' an underlying pattern in the data due to spurious learning of noise or random fluctuations. In the calibration of machine learning models, noise may be unintentionally 'learnt' instead of the actual signal, but this bias may disproportionally worsen in a model with too many parameters compared with available data.

In Bosco et al. [1], we explored an approach to forecast IBCs over 6 months in the future using historical values of IBCs and a selected set of 'drivers' of migration as covariates. This selection came from an extensive literature review of academic publications, including meta-analyses of the literature, as well as already existing online repositories [6]. The study showed the potential of a methodology based on a purely data-driven approach using quantitative data but also highlighted its limitations in constructing consistently accurate predictions over time due to challenges in finding meaningful, timely, and non-redundant signals.

Given these limitations, and in light of the specific policy needs and operational context in which this forecasting model is to be deployed, we opted for a 'mixed' methodological approach, which includes a limited set of covariates plus qualitative assessments from experienced analysts in the migration domain as part of a machine learning architecture to carry on time series forecasting.

To consider the time passed and the seasonality of the time series we used the year and the month of the observation as covariates. In order to give the model the ability to detect that January and December are contiguous months, we converted each month into its sine and cosine values (distributing the 12 months over a unit circle and considering their coordinates). This method, compared with a simple linear indexing of months, avoids to artificially overestimate the time distance between December and January, and in general between the final months of a given year and the initial months of the next year.

To incorporate expert judgement into our methodology we introduced an additional covariate based on the standard deviation of the historical values of the IBCs time series. Specifically, we classified historical IBCs values into three distinct classes *c* based on thresholds defined by the sample standard deviation (*s*) by month of the historical data:

$$c \begin{cases} = 0, & x < s_{month} \\ = 0.5, & s_{month} \leq x < 2s_{month} \\ = 1, & 2s_{month} \leq x \end{cases} \quad (1)$$

where *x* is the number of detected IBCs. Values below one sample standard deviation were assigned to 'class 0', values exceeding two sample standard deviations were assigned to 'class 1', and values in between these two classes were assigned to 'class 0.5' (see Figure 1). This classification, based on absolute magnitude relative to standard deviation thresholds, allows to identify and account for periods of varying magnitude without assuming a normal distribution. This is crucial since the distribution of the realizations in the time series at hand approximately resembles an exponential, or simple monotonic extensions of it (e.g. Weibull distribution). To exemplify considering the case of an exponential distribution, when we split the data in three classes using the thresholds of equation (1), class 0 will contain ~63 % of the observations, class 0.5 about 23 %, and class 1 about 14 %. A more detailed explanation and mathematical proof of this classification is provided in the additional material.

IBC values up to 1 standard deviation represent data points that are relatively low in magnitude compared to the dataset's variability. In contrast, values above 2 standard deviations indicate high-magnitude events, which may correspond to extreme observations. This approach can be classified as an absolute dispersion-based approach, where we categorize data based on their magnitude relative to the sample standard deviation. Many approaches exist for classifying historical time series, and we tested several of these techniques (e.g., Median Absolute Deviation (MAD), K-means, Partitioning Around Medoids (PAM)). However, the simple dispersion-based approach we applied—though slightly different from the more traditional dispersion from the mean—proved to be effective across all routes (see also the additional material).

When new forecast has to be generated, we collect qualitative inputs from the experienced analysts about their expectations on the future developments in the coming months for each migratory route. This qualitative input is summarised and translated into the three quantitative classes which the model uses to predict the future IBCs. We used only three classes to simplify the process for analysts, making it easier for them to assign the correct class for forecasting. This process is discussed in more detail in the 'discussion' section.

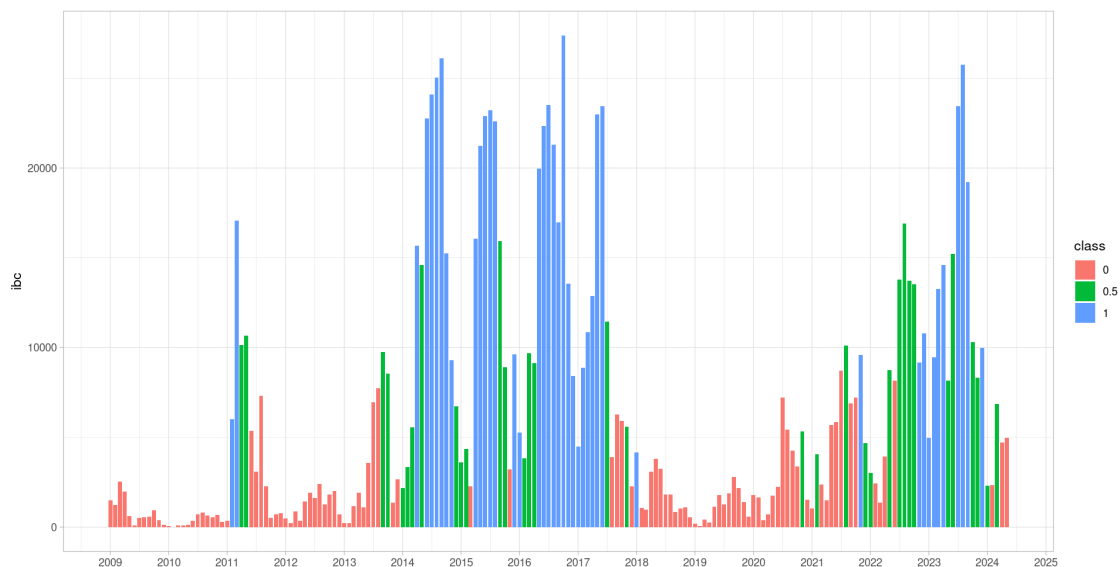

*Figure 1 – Central Mediterranean Route, IBCs classified based on relative magnitude compared to the historical*

The model treats the class values as a continuous variable rather than a categorical one. This means that when predicting future values, any numeric value can be assigned to the class covariate, which makes the model very versatile. This feature is particularly useful when there is uncertainty in choosing among one of the three predefined classes. For example, if the future is expected to be characterized by low arrivals for that route (class 0), but recent trends suggest a slight destabilization, the model allows selecting any intermediate value, such as 0.2. Additionally, it is possible to assign a value beyond the minimum and maximum (i.e. beyond 0 and 1). For instance, if a new shock causes the variable in question to exceed previously observed values, the model can extrapolate beyond the training domain by using a class greater than 1 to make predictions.

## 2.2 Modelling architecture

The machine learning approach we used exploits Artificial Neural Networks (ANNs). The structure of our modelling approach heavily relies on the Semantic Array Programming paradigm (SemAP) as outlined in previous works [7-12]. To address the complexity of modelling and the disparities among input data, parameters, and output, semantic checks were implemented on processed information. Additionally, key components of the model were modularized, aligning with the Semantic Array Programming Paradigm (SemAP) principles, which are largely supported by the Mastrave modeling library [7,9]. SemAP is designed to facilitate computational communication across local contexts, diverse expertise, and disciplines in an uncomplicated yet concise manner. This is accomplished by constraining the versatility of exchanged data using array-based semantic constraints [1,9,13]. In addition, SemAP may also simplify the design and iterative development of non-trivial models, by partly helping to detect semantic inconsistencies.

Semantic Array Programming is utilized to break down a complex data transformation model (D-TM) into distinct logical blocks, making it easier to verify their reliability through the application of mathematical constraints such as precondition, invariant, and postcondition semantic checks. This approach allows even intricate wide-scale transdisciplinary models to be represented as compositions of simpler SemAP blocks.

In the effort of increasing the reproducibility, free scientific software tools and libraries and freely available datasets were used in applying the model, and reproducible techniques were used for applying the different modules that are part of the modelling architecture.

### 2.2.1 Artificial neural networks

An Artificial Neural Network (ANN) is a data-transformation model that generates outputs from given inputs, inspired by the human brain's structure. It consists of interconnected input, intermediate, and output nodes, with connections represented by weighted functions. During learning, the network adjusts these weights by comparing outputs to target values. ANNs can capture both linear and nonlinear relationships, making them effective for handling complex data with noise or weak correlations, as they can identify and exploit intricate relationships between covariates and output data.

A valuable theoretical finding pertains to a specific group of ANNs, namely feed-forward multilayer perceptrons. These ANNs possess the property of universal approximation, signifying that a well-constructed and trained ANN in this family is virtually able to reproduce any relationship between input covariates and the quantity to be modelled [14,15]. Here we applied a feed-forward multilayer perceptron implemented in MATLAB language in GNU Octave. This model was created exploiting the Neural Network Package [16] available in GNU Octave. The hyperparameters subject to tuning encompassed the number of neurons in each layer and the activation function employed in both the hidden and output layers. As a training algorithm, the Levenberg-Marquardt approach was used. It enhances ANNs by efficiently optimizing weights through a combination of gradient descent and Gauss-Newton methods, minimizing overfitting and leading to faster convergence. Its efficient handling of complex optimization makes it more robust in challenging environments and with limited data.

In order to further increase the model's predictive capacity, a simplified version of the Selective Improvement by Evolutionary Variance Extinction (SIEVE) [17] was applied to the ANN. The core of the SIEVE architecture is to iteratively select the best parameter vectors, so reducing exponentially the number of parameter vectors surviving each iteration. This reduction of parameter vectors is typically compensated for through the extension of the computational resources dedicated to training each parameter vector until the optimum vector passes the final SIEVE. The complete (non-simplified) SIEVE architecture includes a "generative" phase (bypassed in the simplified SIEVE) within each step, where a cloud of new vectors is generated close to each surviving parameter vector from the previous sieves [17]. We applied the simplified SIEVE to reduce the required computational time. The simplified version proved to be sufficiently robust to improve model performance [11, 1].

Since the results of an ANN can vary greatly depending on the variability of the random parameter vectors, to make our outcomes more robust we first saved the monthly output of 100 different networks surviving the SIEVE selection and we took the observations that lie between quantiles 10 and 90 to remove possible outliers. We then generated 10,000 artificial monthly series by randomly sampling with replacement the values of each month in each series (bootstrapping). The months in each series were then summed up to provide an aggregated output and the min and max of these were used to compute the range, which is the final output of the model, in line with the requirement of the AMMR. Bootstrap provides a more robust estimate of uncertainty by generating multiple resamples from the original data, offering insights into the variability and distribution of predictions. This method enhances the reliability of the model's uncertainty quantification, reducing potential bias and giving a more comprehensive view of the model's performance compared to the limited perspective offered by quantiles alone. This allows the model to generalize better making its predictions more stable and less sensitive to outliers and anomalies in the data.

2.3 Model performance and validation

In order to maximise model performance, we split the data into training and validation (the last 12 months of data) employing a two-step validation approach that exploit a cross-validation for tuning the model and a validation to measure modelling performance. To retain enough observations for a meaningful validation exercise, we used all samples up to January 2021 included as training set, and the remaining set of the data (February 2021 to August 2024) was used to validate the model. August 2024 was the last available data point at the time we conducted the analysis

With 43 observations per migratory route kept for validation, we created 32 independent and partly overlapping validation sets, each representing a different 12-months period. We started from the period "February 2021 to January 2022" until "September 2023 to August 2024". Each validation set was shifted ahead chronologically by one month from the previous one; for example, given that the first set starts in February 2021, the second one starts in March 2021, the third in April 2021 and so on (see the additional material). For each set we then developed a model trained with data until the first validation point excluded. For example, for the first validation set we used all available observations (since January 2009) until January 2021 included for training the model, whereas for the last one we used all samples until August 2023 included.

Through a repeated random sub-sampling cross-validation applied to the data set for model training, we identified the optimal model hyperparameters. To assess the performance of different modelling architectures, we computed both the root mean square error (RMSE) and the mean absolute error (MAE). While some experts recommend solely relying on MAE for comparing average model performance [18], we opted to include RMSE in our evaluation due to its sensitivity in detecting occasional large predictive errors [19]. Considering the high number of validation sets (32 for each migratory route), and the rather intensive computational process required, we developed a semi-automated approach to validate the model. We kept the same optimal hyperparameters for all validation periods in each route. The tuning of the hyperparameters was therefore done only for the first validation period and then used in all the validation sets.

To assess the model's performance, which reflects the relationship between predicted and observed values, we computed both the mean absolute error (MAE) and the mean absolute percentage error (MAPE). We also included the Precision (i.e. the number of times the real value is within the forecasting range), and the explained variance of the model (expressed in proportional terms) as measures of the model's predictive capacity.

The proposed methodology requires the assignment of a class to each future observation to generate forecast of IBCs. In a real-life forecasting exercise, the exact class of the reference period to be predicted is not known but needs to be assigned based on the experts' input. To simulate this in our validation exercise, we calculated the class values in each validation set using the method described in 2.1.2, and then we 'altered' them to reflect that experts may formulate analytical considerations which can depart from the calculated values. In particular, we split the 12 months' period in each validation set into two equal parts (6 months each), and assigned each the value of the arithmetic mean of the calculated class obtained in the first step. The assignment of only two values reflects more realistically the capability of the analysts to provide an approximate indication of the future class, rather than an exact class for each of the twelve months. In this simulation (which we refer to as Case 1, or "mean class" from now on), we assume that the provided qualitative inputs are more or less adequate: they derive from the calculated monthly classes, but they are averaged to artificially introduce the human bias.

These simulated classes are therefore used to assess the model's performance, whose results are shown in section 3. In the next section we describe a sensitivity analysis which we performed to evaluate how variations in the classification of the future migratory flow would affect the modelling output. In particular, we introduce two additional cases in which the calculated class values are altered in a way different from what is done for Case 1.

2.4 Sensitivity Analysis

Here, we describe the sensitivity analysis we performed to evaluate how variations in the classification of the future migratory flow affect the modelling output. The aim is to assess model robustness and to evaluate the consequences of changing assumptions.

In the previous section we already discussed how we altered the calculated values of the future class covariate using the arithmetic mean (Case 1). Here we introduce two other ways of altering the calculated class values, which we used in our sensitivity analysis. Case 2 ("approximated class") is similar to Case 1, in that we calculated the arithmetic mean of the class for 6-month periods. However, the mean was rounded to the closest default class value as of section 2.1.2 (i.e. either 0 or 'stable', 0.5 or 'moderately unstable', or 1 or 'unstable', see the additional material). Lastly, Case 3 ("precise class") simulates the ideal, though less likely, situation in which the analytical considerations provided by analysts result in assigning the correct class for each of the 12 months of the validation period. In this case, the classes in the validation set are the result of the calculation method described in 2.1.2 (see the additional material).

Considering that the tuning of the hyperparameters was done for the first validation period and applied to all the 32 validation sets, the results of the validation process, even for Case 3 ("precise class") that mirrors the real values, are likely slightly less accurate than if the model had been finely tuned for each validation set separately.

# 3 Results

3.1 Validation results

The following sections shows the performance of the applied modelling architecture using Case 1 ("mean class") and a summary is provided in Table 1. The prediction ranges of Illegal border-crossings for each validation period for Western African Route (figure 2), Central Mediterranean Route (figure 3), Western Mediterranean Route (figure 4), Eastern Mediterranean Route (figure 5) and Western Balkan Route (figure 6) are also presented.

Table 1 shows the error metrics of this validation exercise for each migratory route. In general, the methodology is very robust, as most of the time the true value lies within the forecasting range ("Precision" column in the table), and the errors are generally very low. The error metrics were calculated using the difference between the real value and the value of the forecasting range that is closest to the real value. If the real value was within the forecasting range, this difference amounts to zero, meaning that the model successfully predicted the real value. The prediction range was calculated using the method described in section 2.2.1.

*Table 1: Error metrics and validation results of the forecasting methodology with Case 1 ("mean class"). "Precision" is the number of times the real value was within the forecasting range.*

| Route | Precision (mean class) | Explained Variance | Mean Average Error | Mean Average Percentage Error (%) |
|---|---|---|---|---|
| Western Mediterranean | 30/32 | 0.92 | 89 | 0.5 |
| Central Mediterranean | 22/32 | 0.98 | 2213 | 2 |
| Eastern Mediterranean | 28/32 | 0.98 | 189 | 0.5 |
| Western Balkan | 19/32 | 0.89 | 4936 | 4 |
| Western African | 20/32 | 0.96 | 1570 | 3.5 |

3.1.1 Western African route

Figure 2 presents the model's predictions for illegal border crossings along the Western African Route across 32 distinct validation periods. In 20 out of the 32 periods, the illegal border crossings reported by Frontex on its website fall within the range predicted by the model. Table 1 provides the statistics for this route, showing a precision of over 60 %, high explained variance, and low MAE and MAPE values.

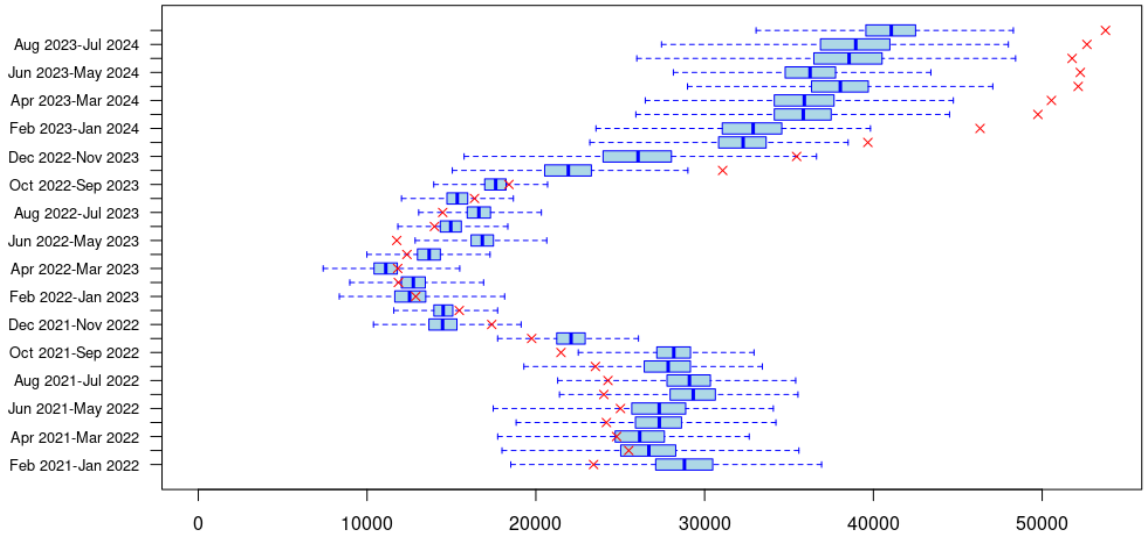

*Figure 2: results of the validation for the Western African Route for 32 validation periods with classes calculated using the Case 1 ("mean class") (forecast in blue, true values in red).*

### 3.1.2 Central Mediterranean Route

Figure 3 illustrates the model's performance in forecasting IBCs along the Central Mediterranean Route over the available validation sets. The actual values fall within the predicted range of the model 22 out of 32 times. Table 1 summarizes the statistics for this route, highlighting a precision close to 70 %, a very high level of explained variance, and low MAE and MAPE values.

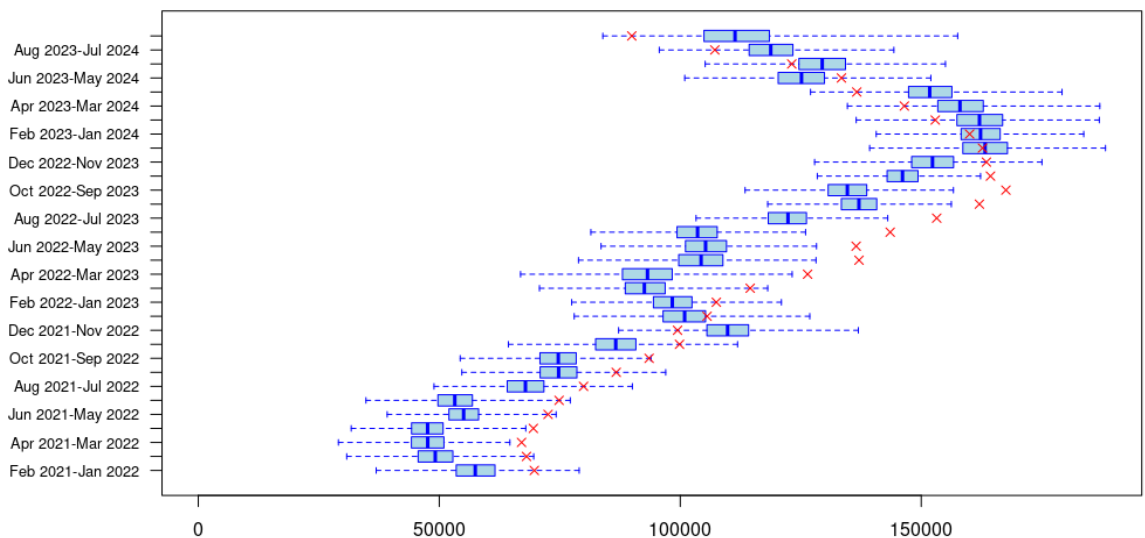

*Figure 3: results of the validation for the Central Mediterranean Route for 32 validation periods with classes calculated using the Case 1 ("mean class")  (forecast in blue, true values in red).*

### 3.1.3 Western Mediterranean Route

Figure 4 presents the model's ability to forecast illegal border crossings along the Western Mediterranean Route across 32 distinct validation periods. In 30 of these periods, the actual IBC figures reported by Frontex on its website fall within the model's predicted range. This demonstrates the model's robustness in capturing the trends of illegal crossings. Table 1 further summarizes key statistical metrics for this route, showcasing the model's high precision of over 87 %, a very high degree of explained variance, and exceptionally low values for MAE and MAPE. These results underscore the model's effectiveness in predicting IBCs over the WMR with a solid level of accuracy.

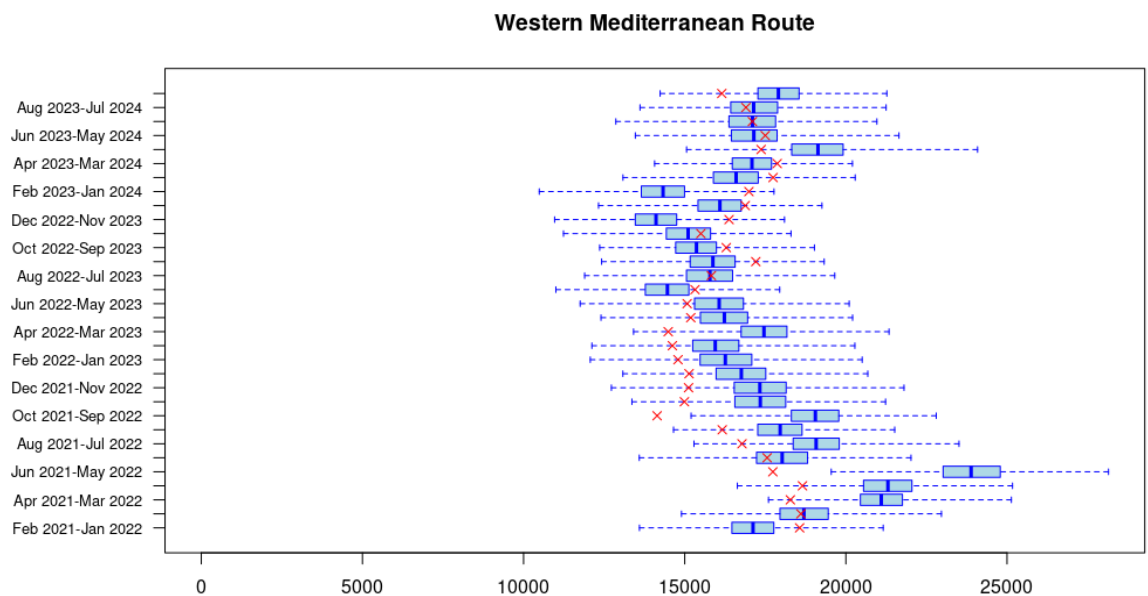

*Figure 4: results of the validation for the Western Mediterranean Route for 32 validation periods with classes calculated using the Case 1 ("mean class") (forecast in blue, true values in red).*

### 3.1.4 Eastern Mediterranean Route

Figure 5 illustrates the model's capacity to predict illegal border crossings along the Eastern Mediterranean Route across 32 separate validation periods. In 28 of these periods, the actual IBC numbers reported by Frontex fall within the range predicted by the model, highlighting its ability to capture trends in illegal crossings effectively. Table 1 provides additional statistical details for this route, demonstrating the model's impressive accuracy with a near 95 % precision, a high level of explained variance, and exceptionally low MAE and MAPE values. These findings emphasize the model's strong performance in forecasting IBCs along the EMR with notable accuracy.

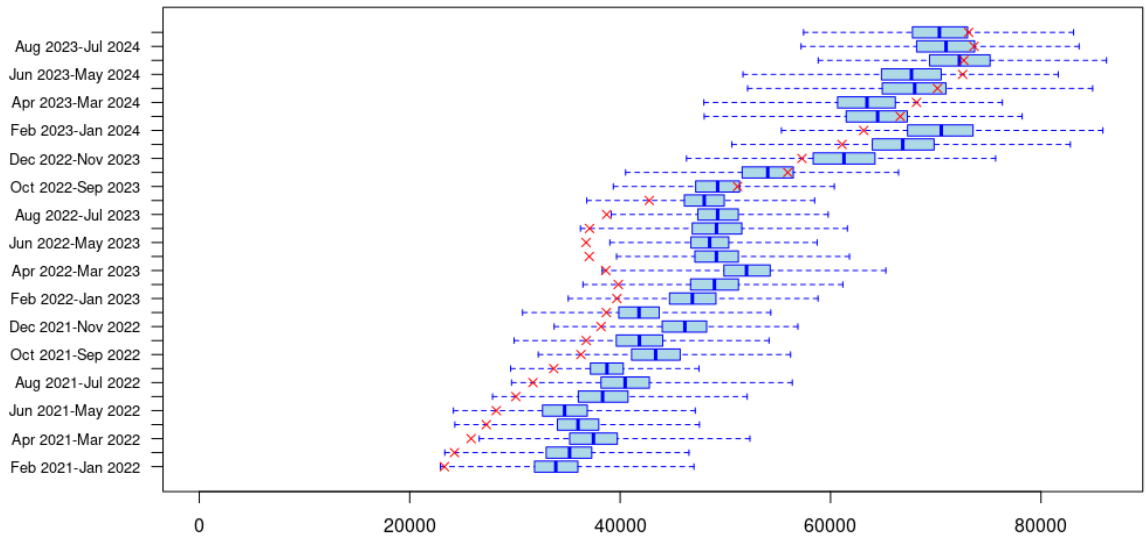

*Figure 5: results of the validation for the Eastern Mediterranean Route for 32 validation periods with classes calculated using the Case 1 ("mean class") (forecast in blue, true values in red).*

3.1.5 Western Balkan Route

Figure 6 displays the model's predictions for IBCs along the Western Balkan Route across the available validation sets. In 19 of these 32 periods, the actual IBC data reported by Frontex aligns within the predicted range of the model, indicating a reasonable level of prediction accuracy. Table 1 further highlights key statistical metrics for this route, revealing a precision rate around 60 %, a high degree of explained variance, and quite low MAE and MAPE values. These results suggest that while the model demonstrates solid predictive capabilities, there may be room for improvement in capturing the full variability of IBC trends for this route.

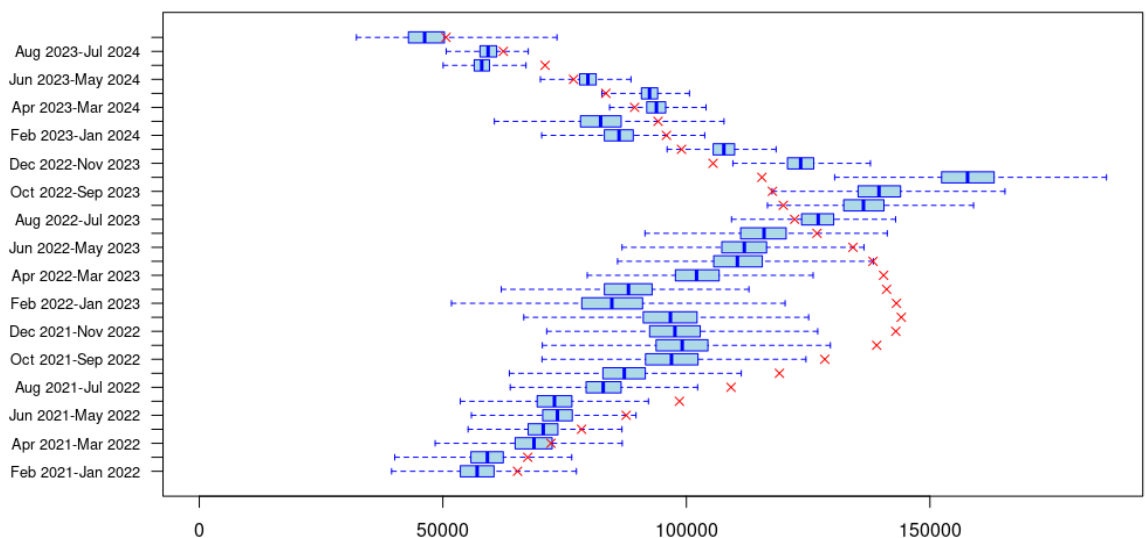

*Figure 6: results of the validation for the Western Balkan Route for 32 validation periods with classes calculated using the Case 1 ("mean class") (forecast in blue, true values in red).*

## 3.2 Sensitivity Analysis

The results described so far are related to the class covariate calculated using the Case 1 ("mean class") described in sections 2.3 and 2.4. Table 2 summarises the results (precision and error metrics) of the three cases used for the sensitivity analysis (see section 2.4). As expected, modelling results get better when using Case 3 (the "precise class"). This was done to test how far off the results of our models would be from an ideal model trained with precise classes (i.e. simulating as if the analysts knew exactly the future developments). We also produced results using the "approximated class", which in theory should provide worst results in terms of precision and error metrics since it approximates the arithmetic mean to the closest default class value (i.e. 0, 0.5 or 1). We indeed observe that overall the "precise class" provides the best results, followed by Case 1 ("mean class") and Case 2 ("approximated class"), as expected. However, even in the worst case, most of the considered routes have still high precisions, high explained variance and low MAE and MAPE.

*Table 2: Sensitivity analysis statistics with error metrics and validation results of the forecasting methodology for the three different cases. "Precision" is the number of times the real value was within the forecasting range.*

| Route/Case | Precision | | | Explained Variance | | | Mean Average Error | | | Mean Average Percentage Error | | |
|---|---|---|---|---|---|---|---|---|---|---|---|---|
| | mean | approx | precise | mean | approx | precise | mean | approx | precise | mean | approx | precise |
| Western Mediterranean | 30/32 | 25/32 | **31/32** | 0.92 | 0.61 | **0.99** | 89 | 262 | **20** | **0.005** | 0.017 | **0.001** |
| Central Mediterranean | 22/32 | 23/32 | **26/32** | 0.98 | 0.97 | **0.99** | 2213 | 2419 | **899** | **0.02** | 0.02 | **0.006** |
| Eastern Mediterranean | 28/32 | 31/32 | **32/32** | 0.98 | 0.99 | **1.00** | 189 | 83 | **0** | **0.005** | 0.003 | **0** |
| Western Balkan | **19/32** | 10/32 | 18/32 | **0.89** | 0.71 | 0.76 | **4936** | 11085 | 7827 | **0.04** | 0.1 | 0.06 |
| Western African | 20/32 | 19/32 | **23/32** | 0.96 | 0.97 | **0.98** | 1570 | 1288 | **1024** | 0.035 | 0.03 | **0.02** |

All the figures of the sensitivity analysis for each migratory route considered can be found in the additional material.

# 4 Discussion

The methodology we presented aims to address the forecasting needs identified within the context of the Migration Pact, including under the framework of the Migration Preparedness and Crisis Blueprint and the Asylum Migration Management Regulation (AMMR) policy cycle. Its key added value lies in the ability to incorporate analytical considerations into forecasting models through the use of the class covariate, representing the qualitative characteristics of each observation in the time series. In particular, the opinion of experienced analysts is used to assign each future value to forecast to one of the three classes described in paragraph 2.1.2. Experts provide an evidence-informed expectation on the future developments in the coming months, based on their assessment of the past and current situation (e.g. in relation to key trends and factors influencing IBCs), on the relevance/impact of key factors shaping those indicators (e.g. ongoing or emerging conflicts, the socio-economic conditions in the main countries of origin and transit, smuggling dynamics as well as anti-smuggling initiatives), and the likelihood of future developments. The analytical considerations are then used to assign a numeric class to the future observations, thus allowing the production of the forecast.

In this work, we tested the performance and robustness of the proposed methodology through an exhaustive validation and a sensitivity analysis, obtaining promising results and an overall error which is rather small and stable. The results presented in Section 3.0 show that the forecasts produced applying this methodology are generally accurate, as most of the time the true value lies within the forecasting range, and the errors (MAE and MAPE) are generally low. However, some limitations related to the proposed methodology should be acknowledged for interpreting these results.

First, the proposed forecasting model relies on historical data as one of its components to produce migration and asylum forecasts. This implies that the approach is subject to the limitations that are common to time series models, in particular when it comes to anticipating events or dynamics that cannot be signalled by past data. To work around this limitation, the methodology presented in this manuscript allows to incorporate analytical considerations from expert analysts about the expected evolution of IBCs based on the most up-to-date analytical evidence available at the time of conducting the forecasting exercise. However, while expert considerations aim to include forward-looking insights identified by EU policy stakeholders (for example, the European External Action Service and the JHA Agencies such as EUAA and Frontex), they still primarily concern past or present dynamics. This implies that the impact of future (unpredictable) developments (political, natural disasters, etc.) may not be considered by analysts when elaborating their considerations.

Despite the overall good results, the validation exercise highlighted that there are differences among routes. For example, the model prediction capacity over the Western Mediterranean and Easter Mediterranean Route is very good. With a very high precision, high or very high explained variance and low MAE and MAPE, the model shows its effectiveness in predicting IBCs over these routes with a solid level of accuracy.

However, the Central Mediterranean Route, the Western Balkan Route and the Western African Route have a relatively lower precisions compared to the other two routes. These routes were more difficult to forecast, for different reasons. Since 2020 the number of detected IBCs in the Western African Route entered a new regime, with an increase of values up to two orders of magnitude compared to the past. Therefore, the number of observations that can help the model to predict these higher values is very limited. In fact, in most cases the values in the forecasting horizon were never observed in the historical data, meaning they are (sometimes much) higher than what the model has been trained with.

In this particular case, even setting the class to 1 (i.e. the highest class in the training set), would lead to an underestimation of the real values in the forecasting horizon. This is what we observe in Figure 3, particularly in the last validation periods (i.e. between "February 2023 to January 2024" and "September 2023 to August 2024"). One way to circumvent this limitation is to assign a class greater than 1 to the observations in the forecasting horizon, which would increase the prediction values allowing the model to extrapolate beyond the training set domain. This is exactly what the modelers can do to simulate a 'new normal'. Similarly, the number of detected IBCs in the Western Balkan Route in recent years fluctuated between high volumes that were the highest since the migration crisis of 2015-16, followed by a decrease in 2023. In 2021 IBCs in this route more than doubled compared to the previous year, whereas in 2022 there was an increase of 136 % compared to 2021. These two years are then followed by a decrease in 2023 that might be attributed to an increase in border controls. Since 2021, indeed, three countries in the Western Balkans – North Macedonia, Serbia, and Albania – host fully-fledged Frontex operations, thereby supporting their authorities in the prevention of future irregular migration to the EU. For these reasons, deriving the classes for the forecasting horizon from the historical values as we did for this validation exercise led to an underestimation of the actual values for the years 2021-22, similar to what we observed for the Western African Route. Finally, the reason for the lower precision in the Central Mediterranean Route might be attributed to the very dynamic political changes in the first half of 2023 in Tunisia, which caused the number of IBCs to reach new highs, as also discussed in Bosco et al. [1].

The sensitivity analysis we conducted provided insights into how sensitive the predictions are to input class changes. Results reported in Table 2 demonstrate the robustness of the model, as the results of the sensitivity analysis show that with moderate changes in the class covariate, very often the real values continue to fall within the model prediction range. This indicates that the model is robust and able to manage moderate variations in the class representing the level of the flow of migrants for each route. Only for the Western Balkan Route the model shows a consistent decrease in its prediction capacity when testing the Case 2 scenario.

This work highlights that producing regular forecasts for IBCs is still a challenge. There are many obstacles, starting from anticipating unexpected environmental, security-related or political shocks, the lack of good quality predictors and the difficulty of working with short historical series of data. Nevertheless, the proposed methodology demonstrated strong performance in predicting IBCs, consistently achieving very low error rates. Even in cases where the actual IBCs value fell outside the model prediction range, the error was never significantly large, indicating that the inclusion of the class covariate improves the model's robustness in making accurate predictions overall. This mixed approach also allows the inclusion of the input from policy analysts to make prediction, which is essential in the operational context the model is used and to make its results more understandable and reliable from a policy perspective.

## Author contributions

C.B. drafted the manuscript. C.B. was responsible for the study design, analysis, and interpretation. U.M. undertook data collection, assembly and produced the datasets. C.B., U.M. and D.d.R. performed data analysis and their technical validation. C.B. and U.M. developed the modelling architecture and developed the code for running the models. J.P. and R.C. followed the policy implications of the study. All the authors contributed to the interpretation and production of the final manuscript. All authors read and approved the final version of the manuscript.

## Competing interests

The author(s) declare no competing interests.

## Data availability

The datasets necessary to support the findings of this article are available in the supplementary material.

# Supplementary material

## S.1. Sensitivity analysis plots

This section provides the results (plots) for the sensitivity analysis by migratory route. This includes the validation using the Case 2 ("approximated") and Case 3 ("precise") classes, for the same 32 validation periods used for "Case 1" class ("mean").

### S.1.1. Western African Route

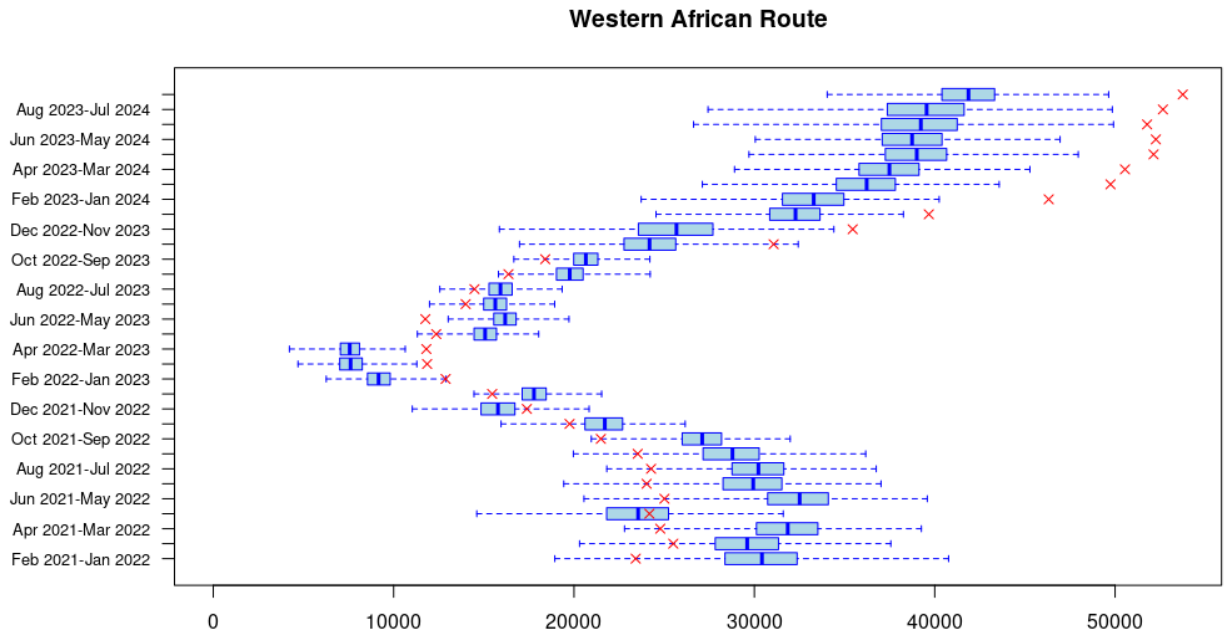

*Figure S.1: results of the validation for the Western African Route for 32 validation periods with classes calculated using the case 2 ("approximated class") (forecast in blue, true values in red).*

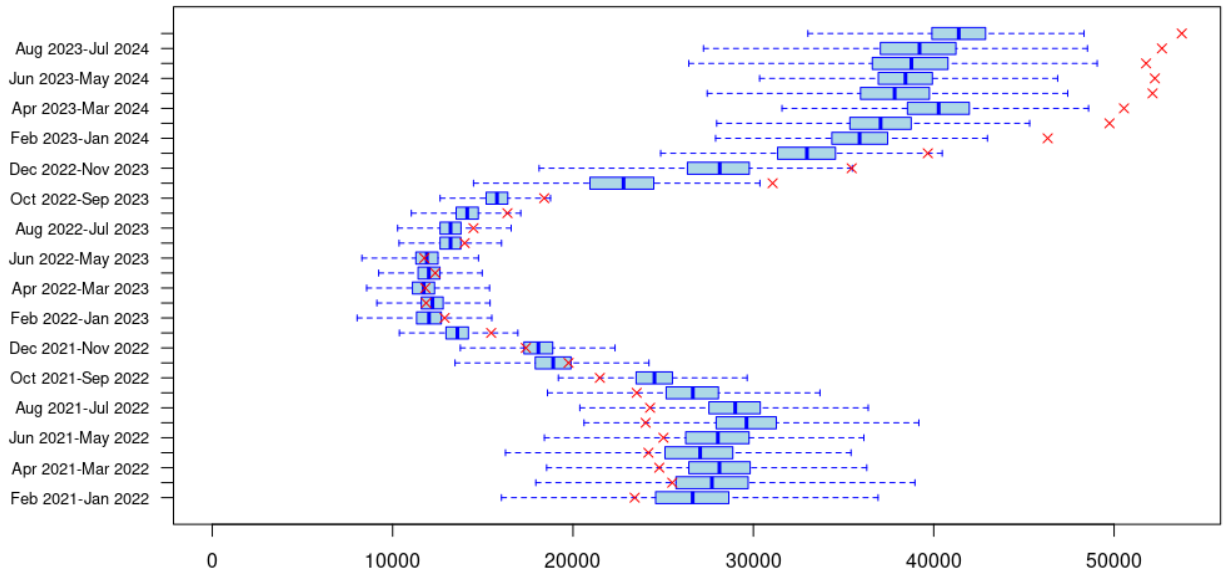

*Figure S.2: results of the validation for the Western African Route for 32 validation periods with classes calculated using the case 3 ("precise class") (forecast in blue, true values in red).*

## S.1.2. Central Mediterranean Route

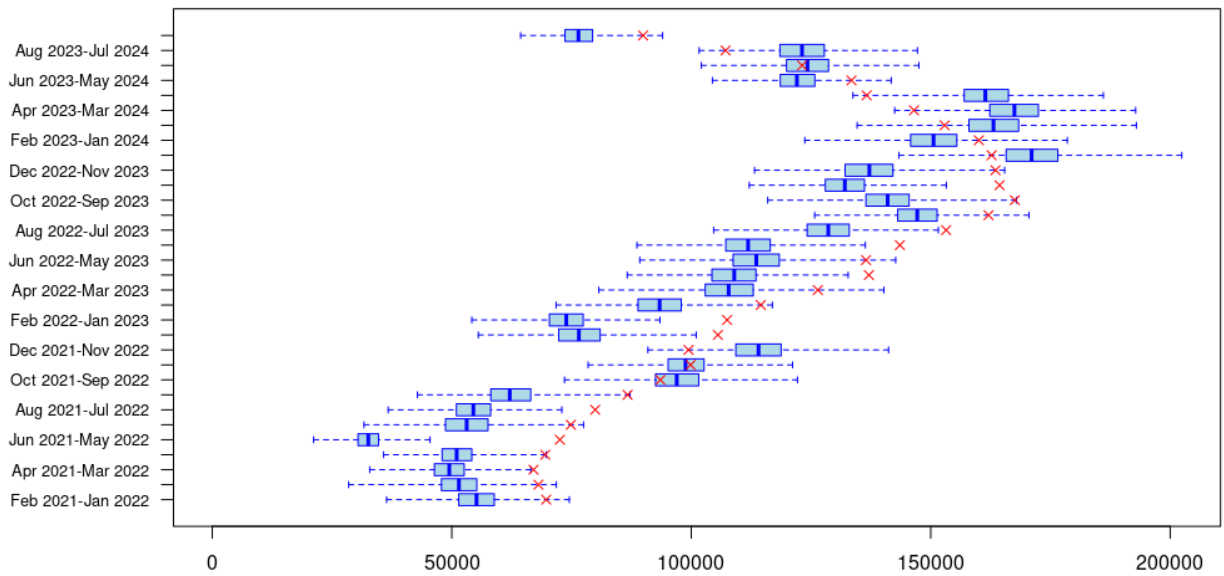

*Figure S.3: results of the validation for the Central Mediterranean Route for 32 validation periods with classes calculated using the case 2 ("approximated class") (forecast in blue, true values in red).*

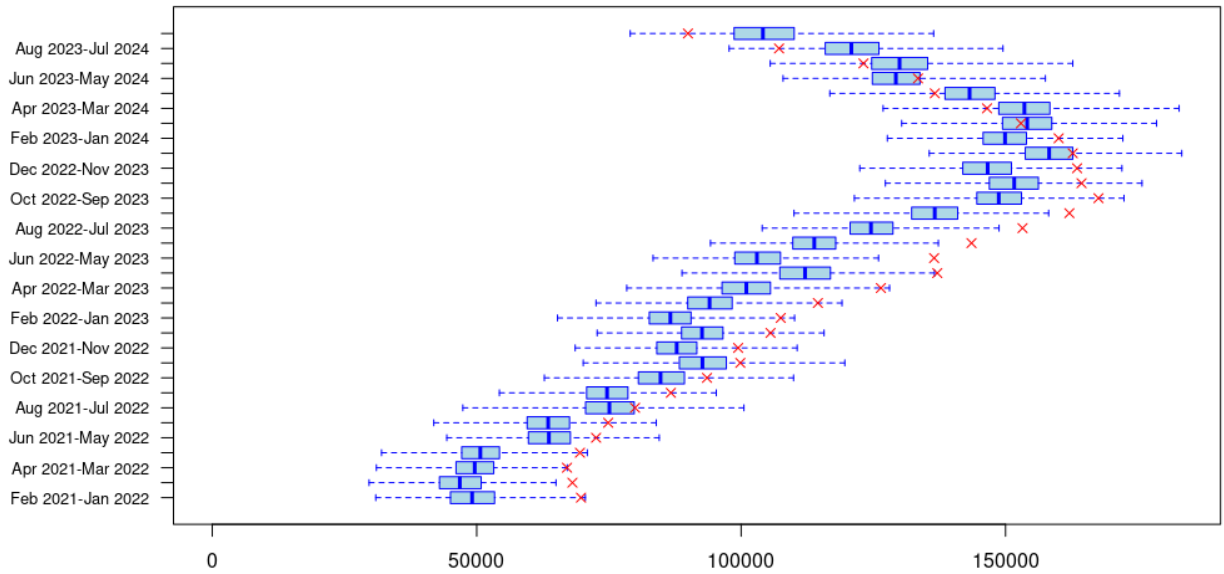

*Figure S.4: results of the validation for the Central Mediterranean Route for 32 validation periods with classes calculated using the case 3 ("precise class") (forecast in blue, true values in red).*

## S.1.3. Western Mediterranean Route

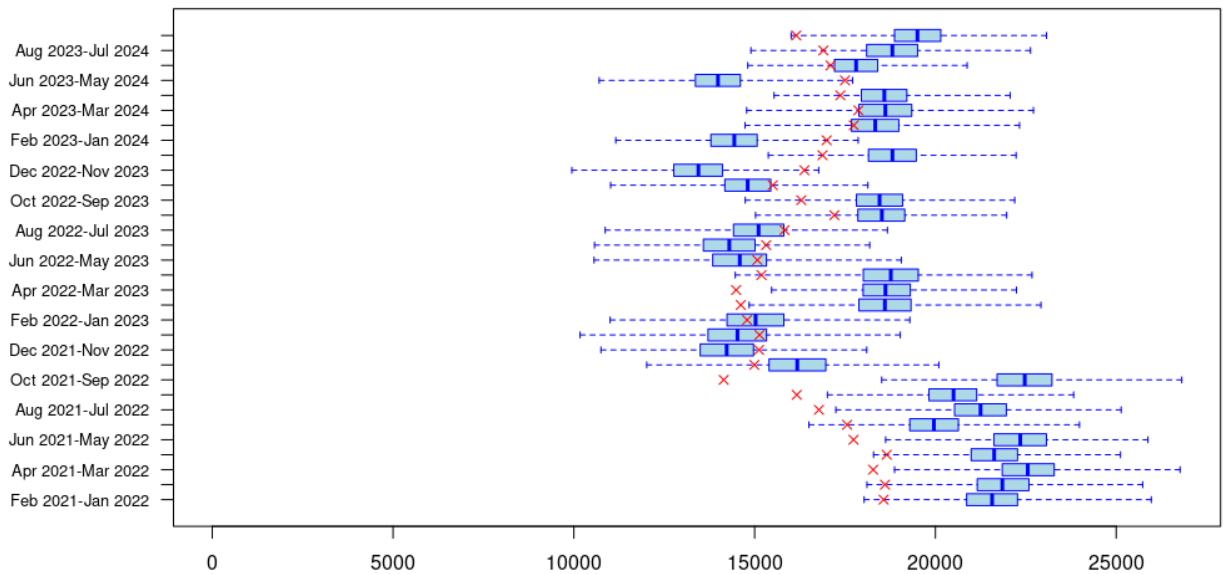

*Figure S.5: results of the validation for the Western Mediterranean Route for 32 validation periods with classes calculated using the case 2 ("approximated class") (forecast in blue, true values in red).*

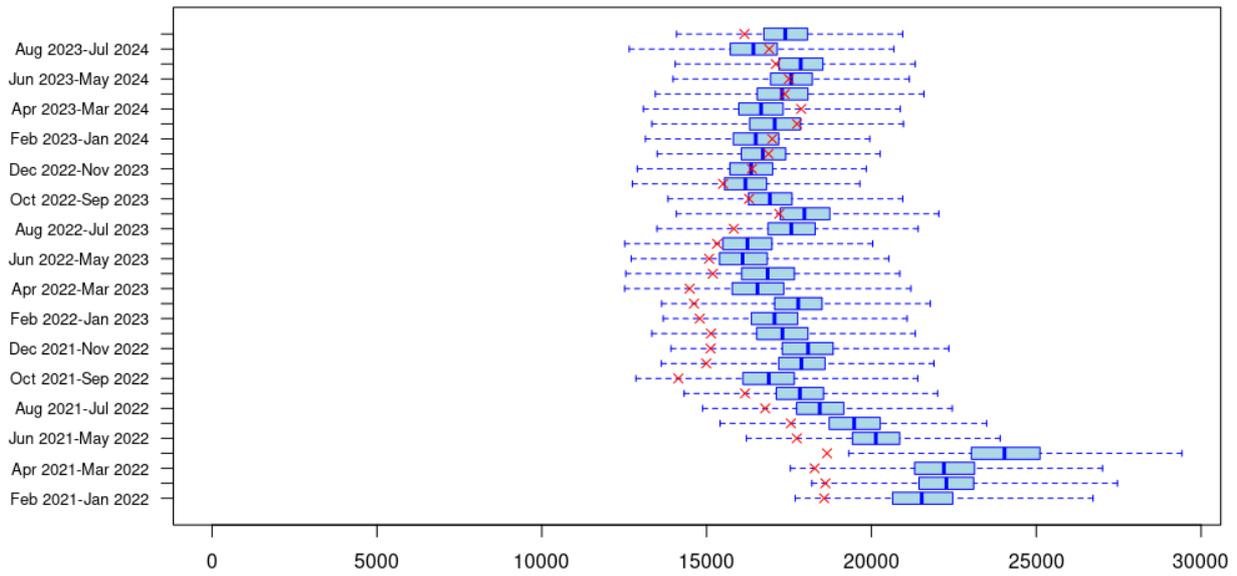

*Figure S.6: results of the validation for the Western Mediterranean Route for 32 validation periods with classes calculated using the case 3 ("precise class") (forecast in blue, true values in red).*

## S.1.4. Eastern Mediterranean Route

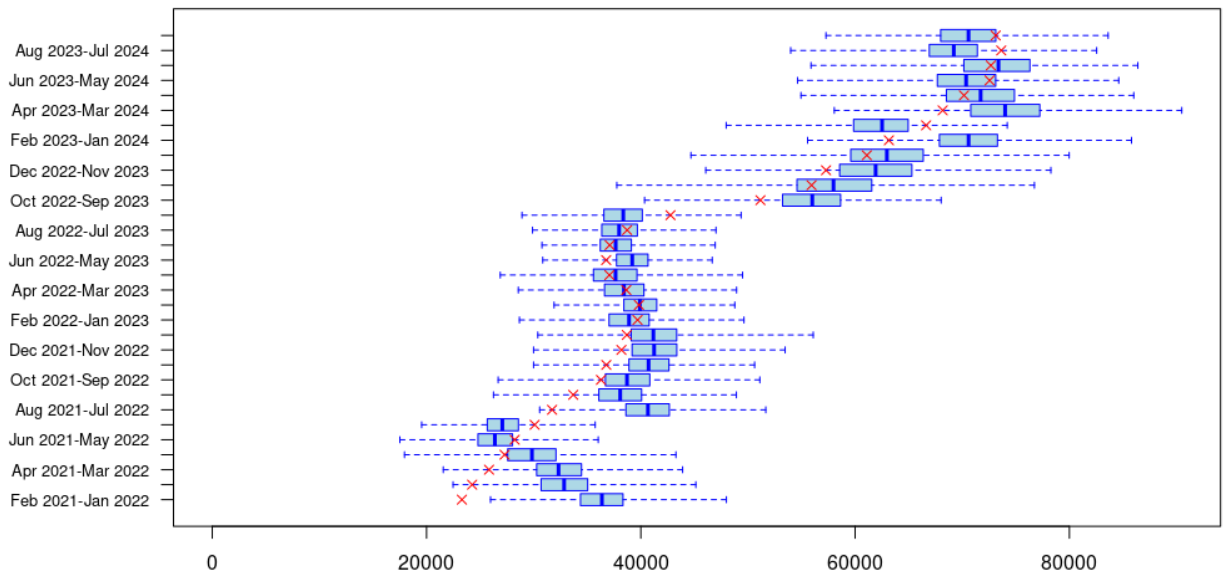

*Figure S.7: results of the validation for the Eastern Mediterranean Route for 32 validation periods with classes calculated using the case 2 ("approximated class") (forecast in blue, true values in red).*

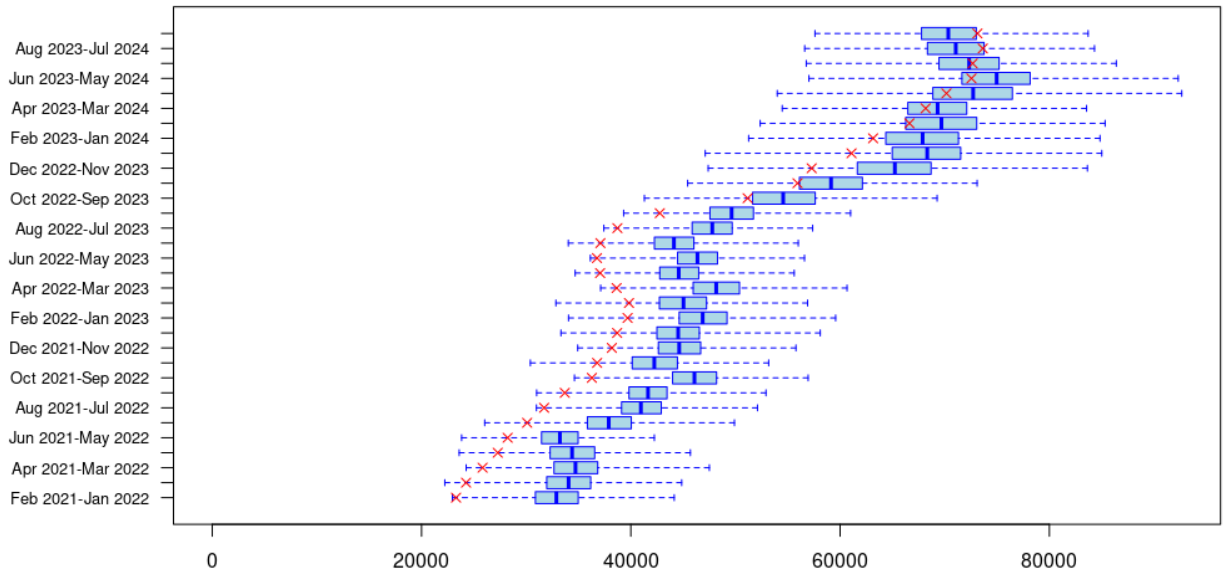

*Figure S.8: results of the validation for the Eastern Mediterranean Route for 32 validation periods with classes calculated using the case 3 ("precise class") (forecast in blue, true values in red).*

## S.1.5. Western Balkan Route

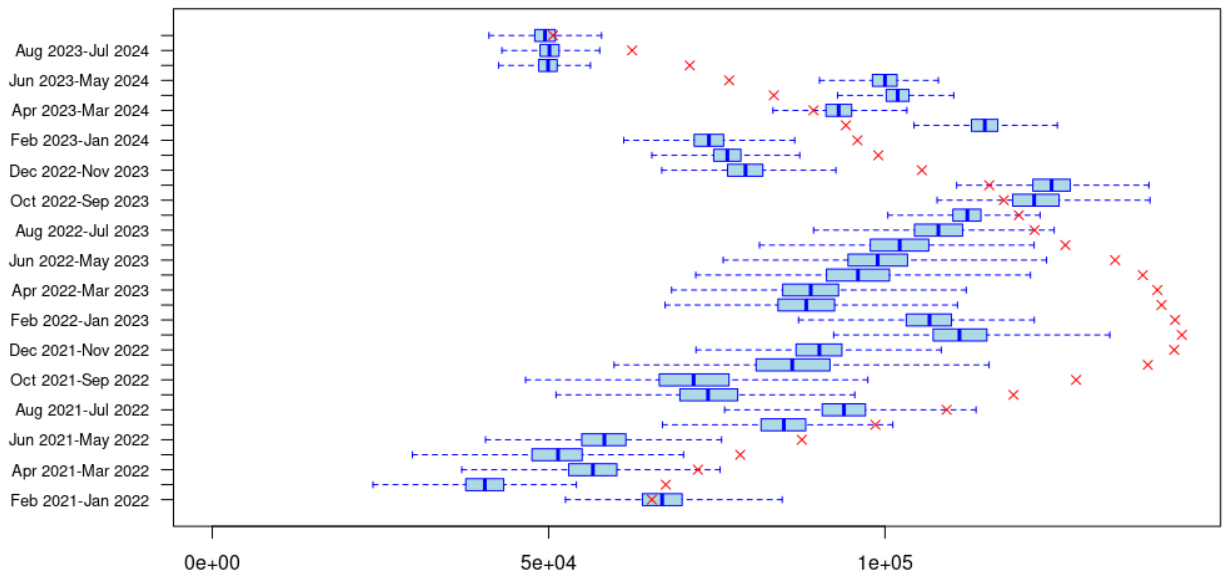

*Figure S.9: results of the validation for the Western Balkan Route for 32 validation periods with classes calculated using the case 2 ("approximated class") (forecast in blue, true values in red).*

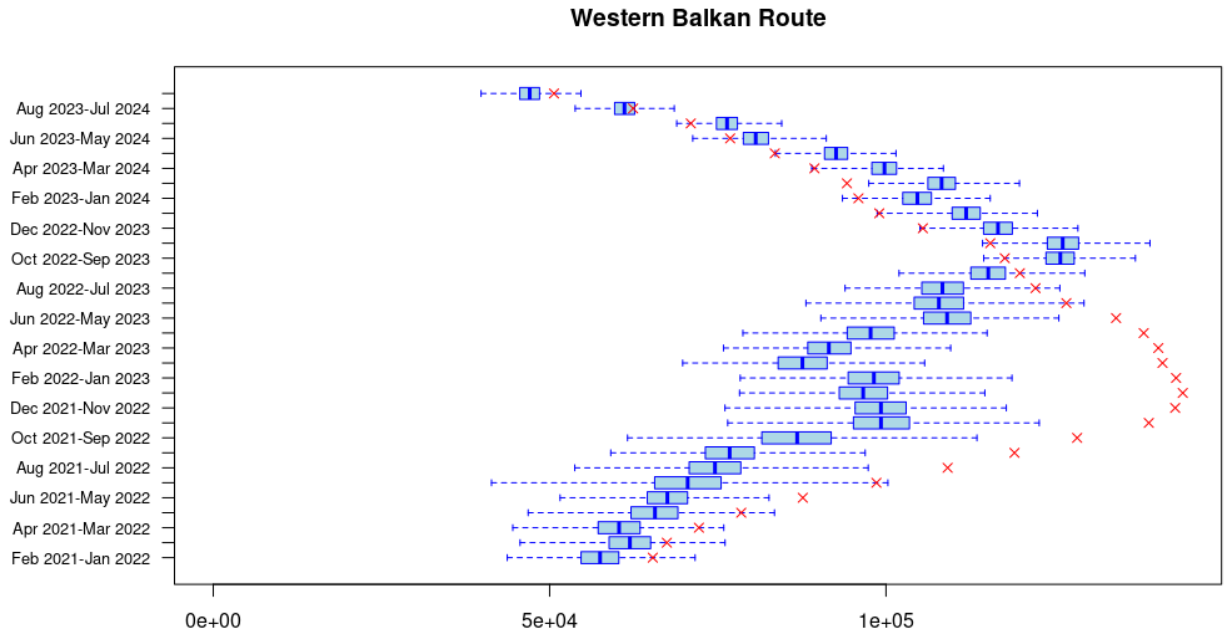

*Figure S.10: results of the validation for the Western Balkan Route for 32 validation periods with classes calculated using the case 3 ("precise class") (forecast in blue, true values in red).*

## S.2. Use of class greater than 1 to evaluate the model performance in case of new maxima

Figure S.11 shows the results of altering the class 1 ("unstable") beyond its default maximum (1) for the Western African Route, to allow the model to extrapolate beyond the domain of the data it was trained with. The scope is to increase the prediction ranges for the last validation set periods, during which the number of detected IBCs was much higher than any historical observation. As can be seen from the figure, the real values in the last validation sets (i.e. from January 2023 on), are almost all within the prediction ranges, thus increasing the Precision from the former 20/32 (allowing for a maximum class of 1), to the current 26/32.

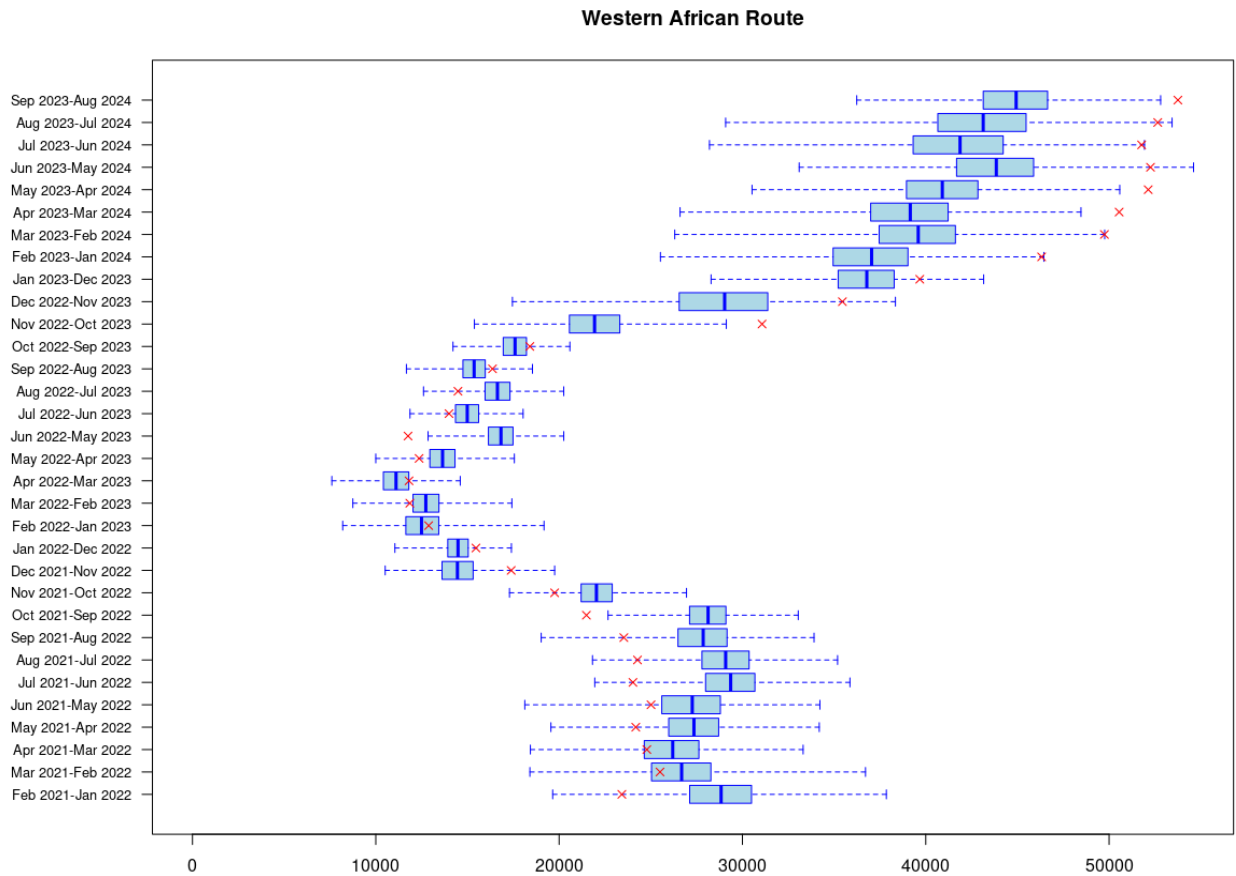

*Figure S.11: results of the validation for the Western African Route for 32 validation periods with classes calculated using the case 1 ("mean class") and a maximum value of 1.2 instead of 1 (forecast in blue, true values in red).*

## S.3. Generation of the 32 validation sets

Figure S.12 is the visual explanation of the method employed to generate the 32 validation sets. Starting from February 2021, the first validation set includes 12 consecutive observations (February 2021 to January 2022). The second validation set starts one month after, namely March 2021, and includes the same number of consecutive observations (12, from March 2021 to February 2022). The same process is repeated until the final available data point is reached (August 2024), and a total of 32 validation sets are generated.

*Figure S.12: the process of generating the (32) validation sets. Orange: observations included in the validation set; light blue: observations not included in the validation set. X axis: available observations that are not part of the training set (43 in total); Y axis: the (32) validation sets.*

## S.4. Alteration of the class covariate

Figure S.13 describes the process that led to the alteration of the class covariate in the three different cases used for the validation and sensitivity exercise (Case 1, 2 and 3). The figure shows an example of the input (on the left), the functions that transformed these input, and an example of the resulting values.

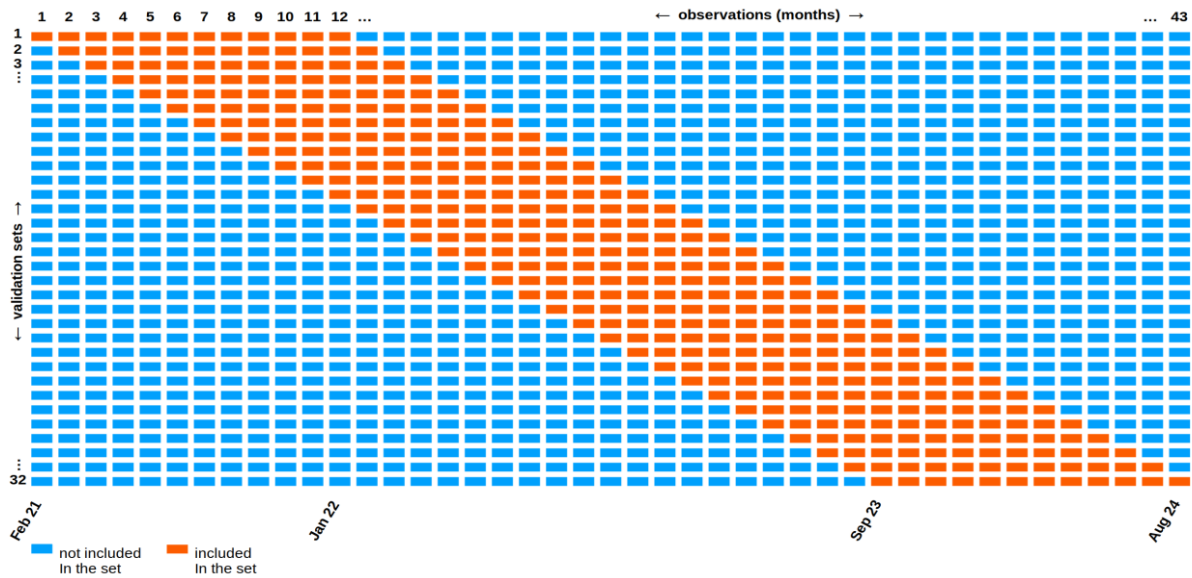

*Figure S.13: process of alteration of the class covariate into the three cases used for the validation and sensitivity analyses (example data).*

## S.5. Dataset List

## S.5.1. Detections of illegal border-crossings statistics (updated monthly)

The dataset of "*Detections of illegal border-crossings statistics (updated monthly)*" used in the manuscript is available at https://www.frontex.europa.eu/what-we-do/monitoring-and-risk-analysis/migratory-map/.

*Table S.1: Data characteristics of Detections of illegal border-crossings statistics (updated monthly)*

| Start date | January 2009 |
|---|---|
| Frequency | Monthly |
| Update frequency | Once every month |
| Publication delay | 2 to 3 months |
| Missing value | No |
| Note | The number of detections of illegal border-crossings do not equate to the number of persons detected, the same person may cross the external border several times or she/he may cross the border without being detected |
| Number of datapoints available at the time of writing | 188 (15 complete years (2009 to 2023) and one incomplete year (Jan-Aug 2024)) |
| Download link | https://www.frontex.europa.eu/what-we-do/monitoring-and-risk-analysis/migratory-map/ |
| Last access | 15$^{th}$ April 2025 |

## S.5.2. Class covariate and definition of the classes

Table S.2 describes the characteristics and definition of the class covariate used as predictor in the mixed model.

*Table S.2: Class covariate characteristics and definition*

| Class | Definition | Class label |
|---|---|---|
| 0 | $X<s$ | stable |
| 0.5 | $s<X<2s$ | moderately unstable |
| 1 | $2s<X$ | unstable |

## S.6. Classification methodology

When the typical range of values should be discriminated against peak occurrences, it may be useful to introduce an intermediate transitional range between low and high values, to account for mixed conditions.

For any random variable X, given its sample mean $m$ and its sample standard deviation $s$, the standard score (also known as z-score) is $Z = (X-m)/s$.

Although for several random variables a partition as simple as

$$[ \ (Z < -1) \quad (-1 \leq Z < 1) \quad (1 \leq Z) \ ]$$
$$\text{lower} \quad \text{intermediate} \quad \text{higher} \quad \text{values}$$
(Equation 1)

may serve to split the data in three intervals to classify lower, intermediate, and higher values, this simplistic method does not work well for random variables with approximately exponential distribution, or simple monotonic extensions of it (e.g. Weibull distribution). For those distributions, it may be useful to consider the reciprocal of the coefficient of variation, also known as signal-to-noise ratio ($SNR = m/s$).[1]

A nonnegative exponentially distributed random variable has $SNR=1$ (because the variance in each exponential distribution is the square of its mean) and the first interval of equation (1) — $Z < -1$ — would be empty ($Z = X/s$ -1 so that $Z < -1$ would require $X < 0$ when the random variable is nonnegative). A similar situation is found for simple monotonic power transformations of an exponential distribution, such as the Weibull distribution, when $SNR < 1$ ($Z = X/s – SNR$, so that $Z < -1$ would require $X$ to be less than a negative quantity). In these cases, a better partition can be obtained by considering two thresholds $a = 1\text{-}SNR$ ; $b = 2\text{-}SNR$. The partition

$$[ \ (Z < a) \quad (a \leq Z < b) \quad (b \leq Z) \ ]$$
$$\text{lower} \quad \text{intermediate} \quad \text{higher} \quad \text{values}$$
(Equation 2)

for an exponential distribution (where $SNR=1$) always splits the values in classes containing respectively ~63% (lower class), ~23% (intermediate class), and ~14% (higher class) of the values. The two thresholds $a$ and $b$ on the standard score $Z$ are equivalent to the simple thresholds

$$[ \ (X < s) \quad (s \leq X < 2s) \quad (2s \leq X) \ ]$$
$$\text{lower} \quad \text{intermediate} \quad \text{higher} \quad \text{values}$$
(Equation 3)

directly applied to the original random variable $X$.

*Proof*

> $a \ < \ Z \ < \ b$
> 
> is equivalent to
> 
> $(1 - SNR) \cdot s \ < \ X - m \ < \ (2 - SNR) \cdot s$
> 
> and to
> 
> $(1 - m/s) \cdot s + m \ < \ X \ < \ (2 - m/s) \cdot s + m$
> 
> which is the same as
> 
> $s \ < \ X \ < \ 2 \cdot s$

---

[1] For *SNR*, also other alternative definitions exist in the literature.

## S.6.1. *Empirical vs. Weibull cumulative distribution function*

The figures in this section highlight the differences between the Empirical CDF and the Weibull CDF by month per each migratory route, in function of the (natural logarithm of the) detected illegal border crossings.

For each route, acronyms are used:
- WAR: Western African Route
- CMR: Central Mediterranean Route
- WMR: Western Meridional Route
- EMR: Eastern Meridional Route
- WBR: Western Balkan Route

Although approximated, the simplified Weibull distribution associated to each route and month is able to show the monthly dynamics due to cyclostationary factors which may affect the observations.

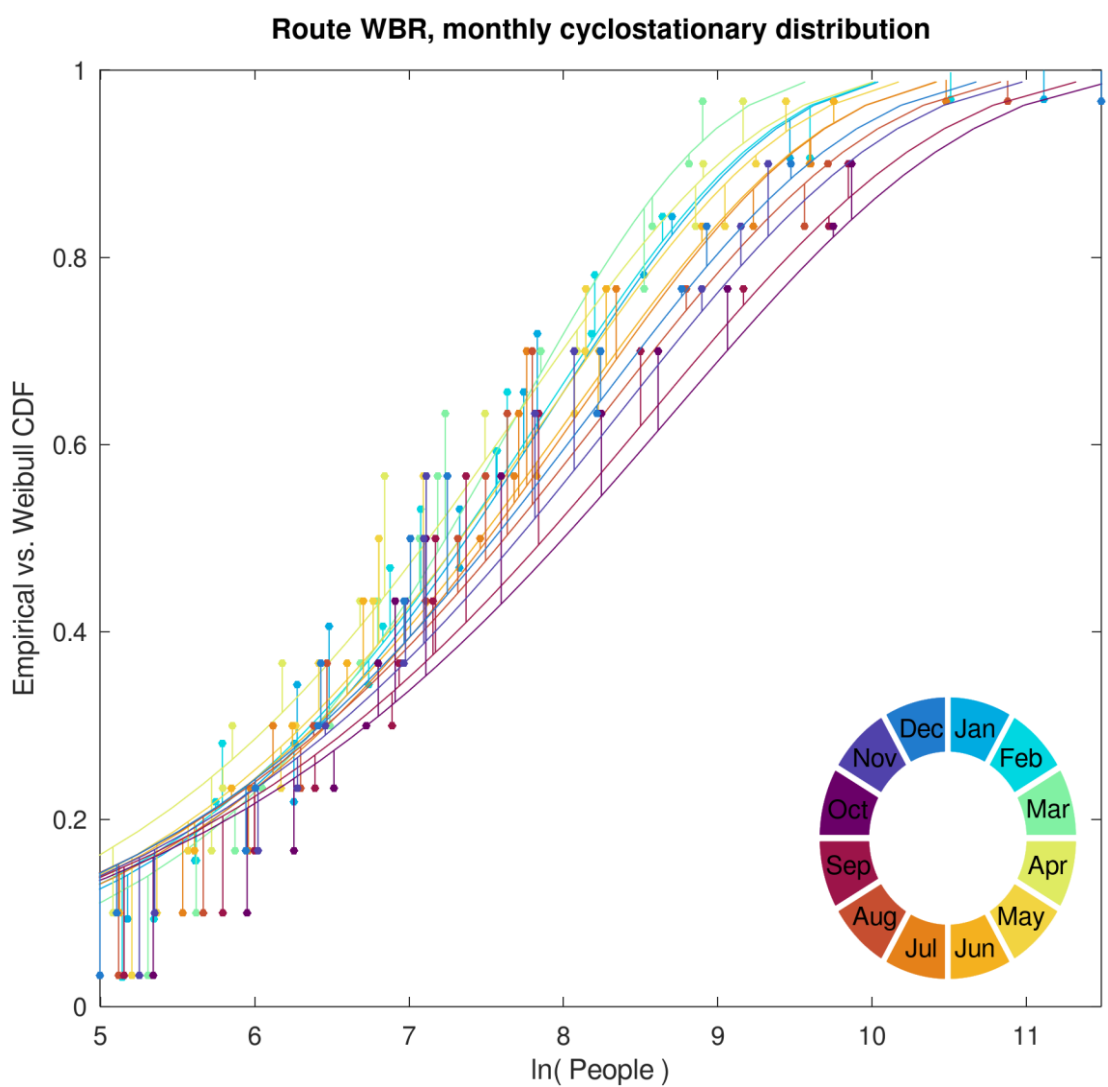

*Figure S.14: Empirical vs. Weibull cumulative distribution function by month for the Western Balkan Route.*

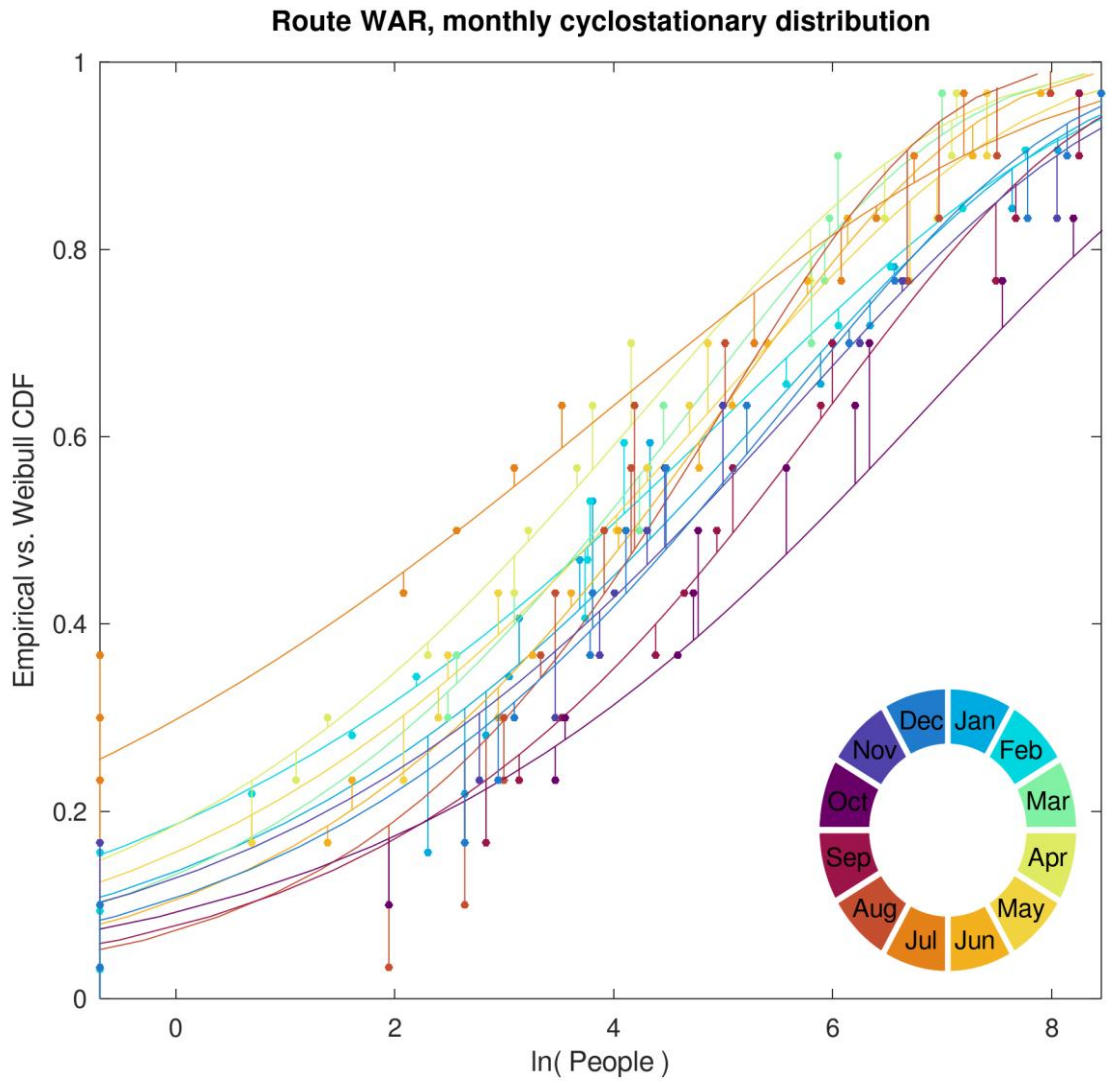

*Figure S.15: Empirical vs. Weibull cumulative distribution function by month for the Western African Route.*

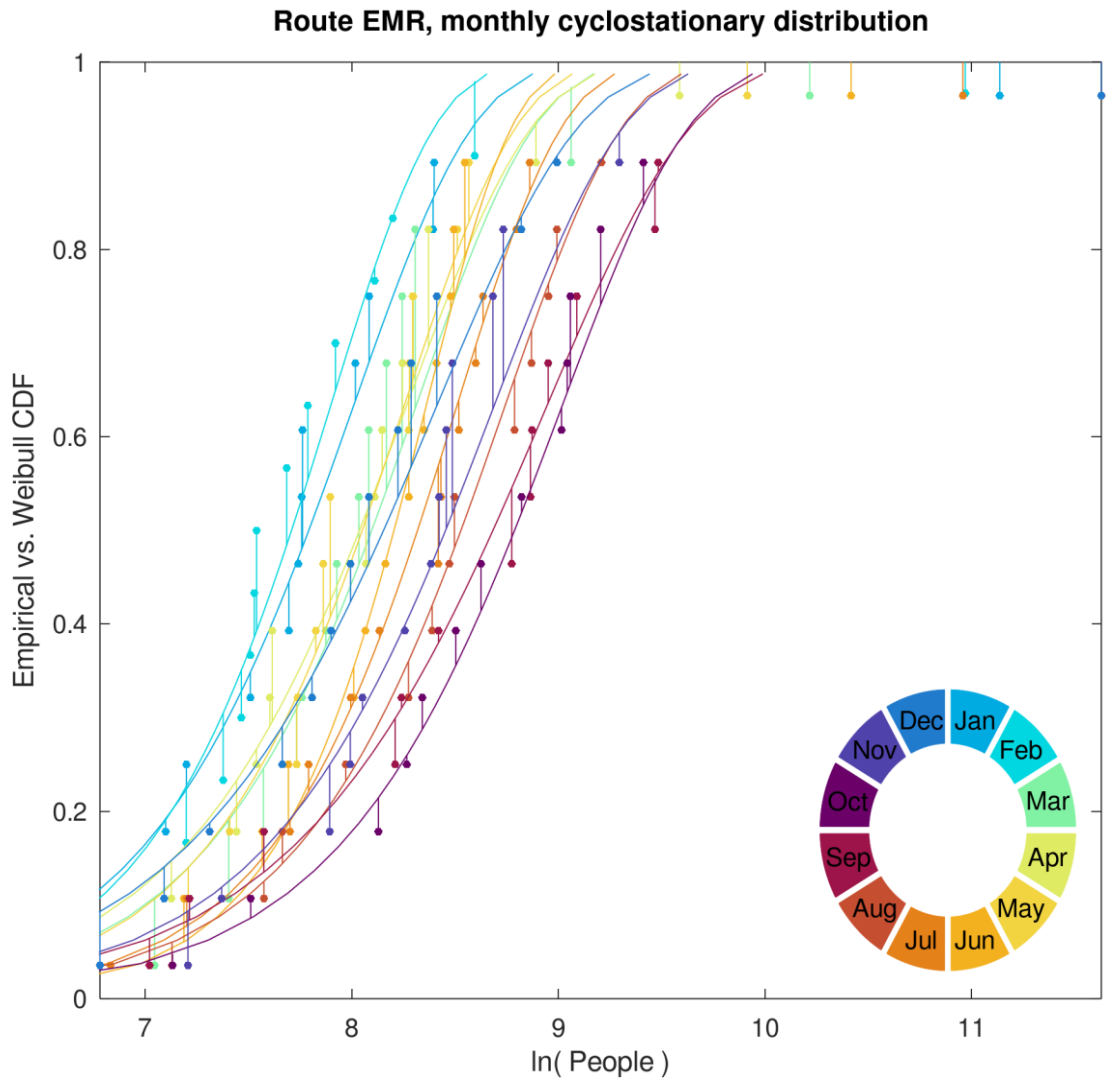

Figure S.16: Empirical vs. Weibull cumulative distribution function by month for the Eastern Mediterranean Route.

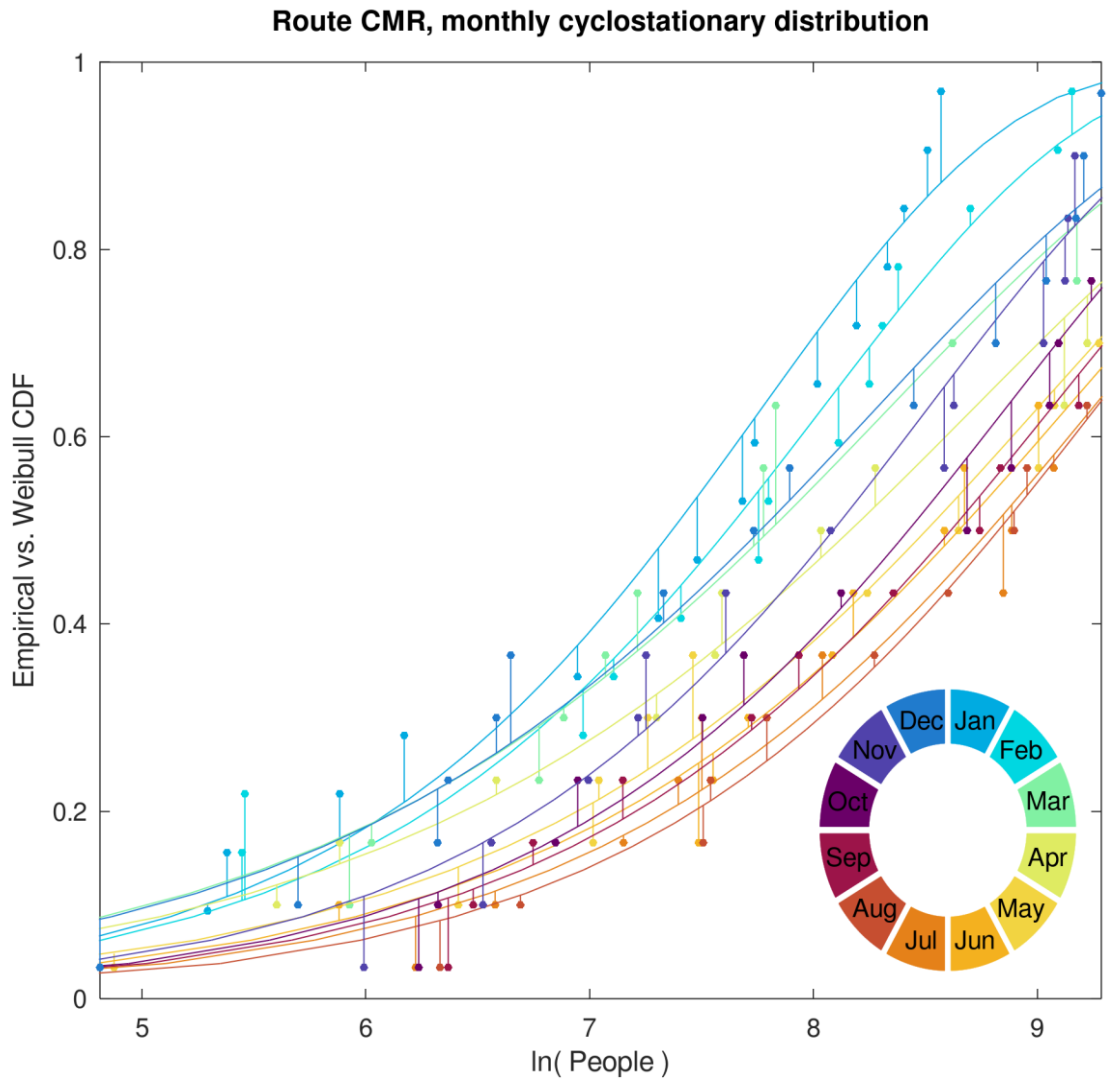

*Figure S.17: Empirical vs. Weibull cumulative distribution function by month for the Central Mediterranean Route.*

## S.6.2. *Empirical vs. Weibull frequency distribution by class*

The figures in this section compare the empirical partition in classes by route and month, versus the corresponding partition in classes which would have occurred under a Weibull distribution.

For each route and month, the signal-noise ratio *SNR* is computed, and the frequency of points in each class is annotated. The progression of classes from the 1st lower class to the 3rd higher class is illustrated in the figures.

| *Progression* | *Values* | *Nominal covariate value associated to each class* |
|---|---|---|
| 1st class | Lower values | 0 |
| 2nd class | Intermediate values | 0.5 |
| 3rd class | Higher values | 1 |

In Table S.2 the signal-noise ratio (SNR) and the frequency of each class are shown for the stationary case (all data undifferentiated, i.e. irrespective of the month), and for the cyclostationary case (data separately assessed by month). It may be noted how the distribution of each route's values by class differs, with a more useful partition in the cyclostationary case, able to better discriminate the higher values, and to detect values which otherwise (in the stationary case) would have been classified as lower values.

In Fig. S.18, per each month and route the frequency of values in the class "0" is shown (vertical axis) along with the corresponding signal-noise ratio (SNR, horizontal axis). The CMR route shows SNR values near 1, while WAR displays lower SNR values, not very dissimilar in different months. The WBR and EMR routes are also characterised by lower SNR values, but their variability by month is noticeably higher. Finally, the WMR route displays a wide range of SNR values in different months, with the highest SNR values found in all the routes considered. In general, lower signal-noise ratios may be associated with a higher frequency of the class "0", while conversely higher SNR values can be associated with more frequent peaks and instability, so that the frequency of class "0" tends to be lower.

In Fig. S.19 and S.20, per each month and route the frequency of values respectively in the class "0.5" (Fig. S.19) and "1" (Fig S.20) is shown along with the corresponding signal-noise ratio (SRN).

In all the Figures S.18, S.19, S.20, the empirical frequency in each class (respectively, class "0", "0.5", "1") is compared with the frequency which would be observed assuming a simplified approximated distribution in each month and route (Weibull distribution) rather than the unknown actual distribution.

Given that the number of values **N** per each month is small (see Table S.2), the uncertainty for the simplified case following a Weibull distribution is noticeable. To account for this small-sample-size uncertainty, a statistical Monte Carlo simulation with **N** defined as for the available data in each route and month was repeated 10000 times, using for each repetition the same Weibull parameters – which are defined as the ones best approximating each month for each route (see S.6.1). The Weibull distribution is continuous, while the number of people in each route is of course a series of nonnegative integer values: to account for this, each Monte Carlo simulated value was rounded to the nearest integer. In the Figures S.18, S.19, S.20, 90 % of the Monte Carlo uncertainty is shown (darker vertical grey bars: range between quantile 5 % and quantile 95 % of the 10000 Monte Carlo runs for each route and month).

For comparison, another set of Weibull simulations was computed, with **N** = 10000 (so, much higher than the limited available data for each route). Again, 10000 Monte Carlo simulations were run (continuous values, without rounding to integer), and their associated uncertainty was represented as before: 90 % of the uncertainty (range between quantile 5 % and quantile 95 %) with 300 logarithmically equi-spaced values of SNR, illustrated with paler vertical grey bars. The reader may compare this reference Weibull simulation with the values mentioned in the section 2.1.2 of the main article, which considers the case of SNR = 1 (where the Weibull distribution is equivalent to an exponential distribution).

Table S.2: Statistics per each route. The signal-noise ratio (SNR) and the frequency of each class are shown for the stationary case (all data undifferentiated), and for the cyclostationary case (data by month). The standard deviation is denoted as **s**. The sample size (whole route, by class, or by month) is denoted as **N**.

| | Stationary: whole year | | | Cyclostationary: statistics for the data by month | | | | | | | | | | | | | |
|---|---|---|---|---|---|---|---|---|---|---|---|---|---|---|---|---|---|
| | | | | Jan | Feb | Mar | Apr | May | Jun | Jul | Aug | Sep | Oct | Nov | Dec | | |
| **Central Mediterranean Route (CMR)** | | | | | | | | | | | | | | | | | |
| mean/stdev (SNR) | 0.911 | | | 1.22 | 1.04 | 0.843 | 0.995 | 1.01 | 1 | 1.09 | 1.09 | 1.05 | 0.971 | 1.13 | 0.993 | | |
| | Freq. by class | N 182 | Freq. by month → | 16 % | 16 % | 15 % | 15 % | 15 % | 15 % | 15 % | 15 % | 15 % | 15 % | 15 % | 15 % | N 182 | Freq. by class |
| Class "0" (lower) ∈ [0 s) | 65.4 % | 119 | | 50 % | 56.2 % | 66.7 % | 60 % | 53.3 % | 66.7 % | 60 % | 60 % | 60 % | 60 % | 53.3 % | 60 % | 107 | 58.8 % |
| Class "0.5" (interm.) ∈ [s 2s) | 19.2 % | 35 | | 25 % | 25 % | 20 % | 13.3 % | 26.7 % | 6.67 % | 13.3 % | 13.3 % | 20 % | 26.7 % | 20 % | 13.3 % | 34 | 18.7 % |
| Class "1" (higher) ∈ [2s max] | 15.4 % | 28 | | 25 % | 18.8 % | 13.3 % | 26.7 % | 20 % | 26.7 % | 26.7 % | 26.7 % | 20 % | 13.3 % | 26.7 % | 26.7 % | 41 | 22.5 % |
| **Eastern Meridional Route (EMR)** | | | | | | | | | | | | | | | | | |
| mean/stdev (SNR) | 0.36 | | | 0.398 | 0.414 | 0.73 | 1.07 | 0.858 | 0.672 | 0.544 | 0.442 | 0.425 | 0.375 | 0.375 | 0.383 | | |
| | Freq. by class | N 169 | Freq. by month → | 14 % | 15 % | 14 % | 14 % | 14 % | 14 % | 14 % | 14 % | 14 % | 14 % | 14 % | 14 % | N 169 | Freq. by class |
| Class "0" (lower) ∈ [0 s) | 94.1 % | 159 | | 92.9 % | 93.3 % | 85.7 % | 64.3 % | 78.6 % | 92.9 % | 92.9 % | 92.9 % | 92.9 % | 92.9 % | 92.9 % | 92.9 % | 150 | 88.8 % |
| Class "0.5" (interm.) ∈ [s 2s) | 1.18 % | 2 | | 0 % | 0 % | 7.14 % | 21.4 % | 14.3 % | 0 % | 0 % | 0 % | 0 % | 0 % | 0 % | 0 % | 6 | 3.55 % |
| Class "1" (higher) ∈ [2s max] | 4.73 % | 8 | | 7.14 % | 6.67 % | 7.14 % | 14.3 % | 7.14 % | 7.14 % | 7.14 % | 7.14 % | 7.14 % | 7.14 % | 7.14 % | 7.14 % | 13 | 7.69 % |
| **Western African Route (WAR)** | | | | | | | | | | | | | | | | | |
| mean/stdev (SNR) | 0.409 | | | 0.47 | 0.491 | 0.648 | 0.557 | 0.613 | 0.501 | 0.577 | 0.548 | 0.628 | 0.469 | 0.501 | 0.551 | | |
| | Freq. by class | N 182 | Freq. by month → | 16 % | 16 % | 15 % | 15 % | 15 % | 15 % | 15 % | 15 % | 15 % | 15 % | 15 % | 15 % | N 182 | Freq. by class |
| Class "0" (lower) ∈ [0 s) | 87.9 % | 160 | | 81.2 % | 81.2 % | 66.7 % | 80 % | 73.3 % | 86.7 % | 73.3 % | 80 % | 73.3 % | 86.7 % | 80 % | 80 % | 143 | 78.6 % |
| Class "0.5" (interm.) ∈ [s 2s) | 6.04 % | 11 | | 12.5 % | 12.5 % | 26.7 % | 6.67 % | 13.3 % | 6.67 % | 13.3 % | 6.67 % | 13.3 % | 6.67 % | 6.67 % | 6.67 % | 20 | 11 % |
| Class "1" (higher) ∈ [2s max] | 6.04 % | 11 | | 6.25 % | 6.25 % | 6.67 % | 13.3 % | 13.3 % | 6.67 % | 13.3 % | 13.3 % | 13.3 % | 6.67 % | 13.3 % | 13.3 % | 19 | 10.4 % |
| **Western Balkan Route (WBR)** | | | | | | | | | | | | | | | | | |
| mean/stdev (SNR) | 0.308 | | | 0.393 | 0.495 | 0.92 | 0.802 | 0.769 | 0.716 | 0.546 | 0.494 | 0.374 | 0.341 | 0.326 | 0.372 | | |
| | Freq. by class | N 182 | Freq. by month → | 16 % | 16 % | 15 % | 15 % | 15 % | 15 % | 15 % | 15 % | 15 % | 15 % | 15 % | 15 % | N 182 | Freq. by class |
| Class "0" (lower) ∈ [0 s) | 95.6 % | 174 | | 93.8 % | 87.5 % | 66.7 % | 66.7 % | 80 % | 80 % | 80 % | 80 % | 93.3 % | 93.3 % | 93.3 % | 93.3 % | 153 | 84.1 % |
| Class "0.5" (interm.) ∈ [s 2s) | 1.1 % | 2 | | 0 % | 6.25 % | 6.67 % | 13.3 % | 0 % | 6.67 % | 13.3 % | 13.3 % | 0 % | 0 % | 0 % | 0 % | 9 | 4.95 % |
| Class "1" (higher) ∈ [2s max] | 3.3 % | 6 | | 6.25 % | 6.25 % | 26.7 % | 20 % | 20 % | 13.3 % | 6.67 % | 6.67 % | 6.67 % | 6.67 % | 6.67 % | 6.67 % | 20 | 11 % |
| **Western Meridional Route (WMR)** | | | | | | | | | | | | | | | | | |
| mean/stdev (SNR) | 0.888 | | | 1 | 1.41 | 2.06 | 1.54 | 1.27 | 0.912 | 0.865 | 1.21 | 0.976 | 0.987 | 0.906 | 1.21 | | |
| | Freq. by class | N 182 | Freq. by month → | 16 % | 16 % | 15 % | 15 % | 15 % | 15 % | 15 % | 15 % | 15 % | 15 % | 15 % | 15 % | N 182 | Freq. by class |
| Class "0" (lower) ∈ [0 s) | 75.3 % | 137 | | 62.5 % | 43.8 % | 13.3 % | 40 % | 53.3 % | 80 % | 66.7 % | 53.3 % | 66.7 % | 66.7 % | 73.3 % | 53.3 % | 102 | 56 % |
| Class "0.5" (interm.) ∈ [s 2s) | 16.5 % | 30 | | 31.2 % | 25 % | 40 % | 33.3 % | 33.3 % | 13.3 % | 26.7 % | 40 % | 20 % | 26.7 % | 13.3 % | 33.3 % | 51 | 28 % |
| Class "1" (higher) ∈ [2s max] | 8.24 % | 15 | | 6.25 % | 31.2 % | 46.7 % | 26.7 % | 13.3 % | 6.67 % | 6.67 % | 6.67 % | 13.3 % | 6.67 % | 13.3 % | 13.3 % | 29 | 15.9 % |

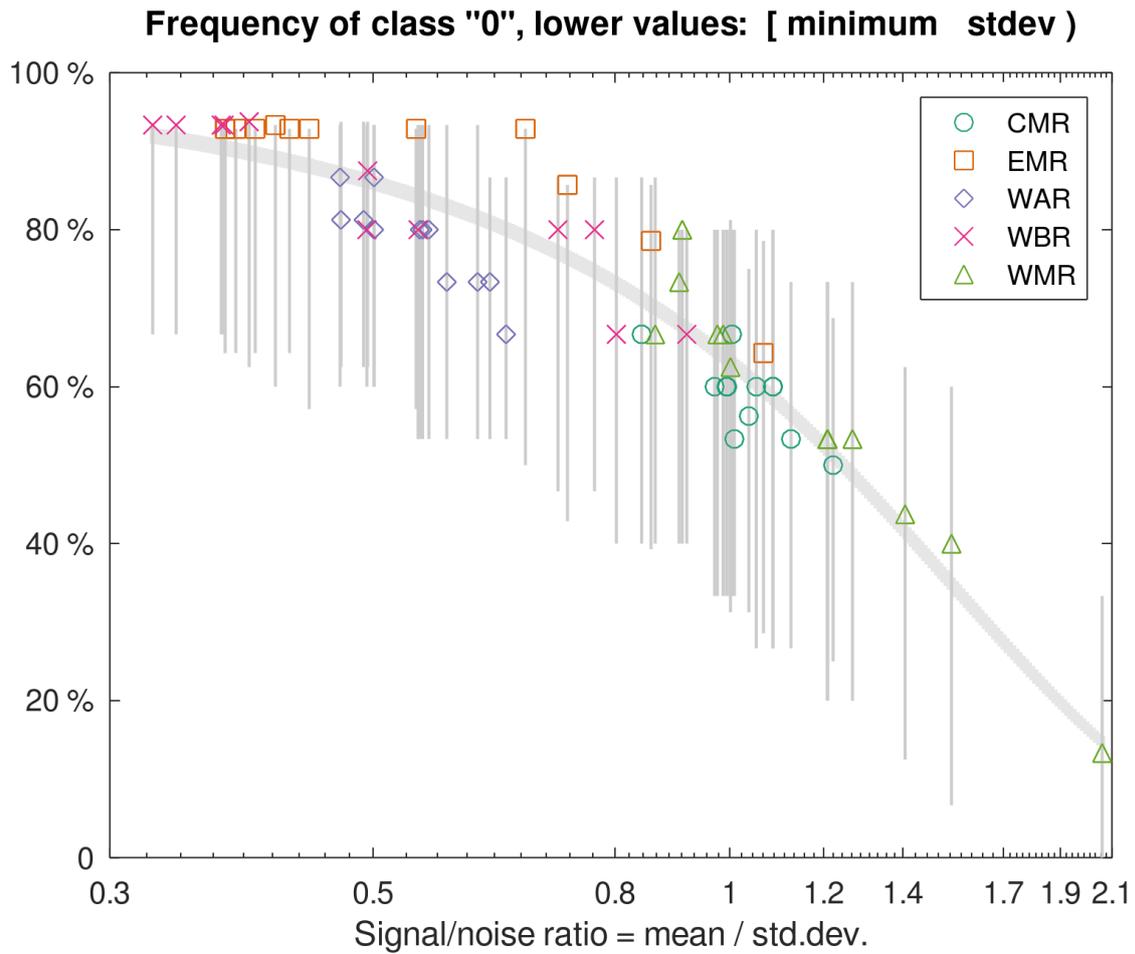

*Figure S.18: frequency of values in the class "0" per each month and route (vertical axis) along with the corresponding signal-noise ratio (SNR, horizontal axis). The uncertainty of 10000 Monte Carlo simulations using a simplified approximation (Weibull distribution) for the unknown actual distribution is represented with the dark-grey vertical bars associated to each point (90 % of the total uncertainty, range between quantiles 5 % and 95 %, after rounding the simulated values to their nearest integer). The pale-grey vertical bars show a reference Weibull distribution (N=10000 instead of the much smaller sample size of each route in each month). Again, the 90 % uncertainty range is represented.*

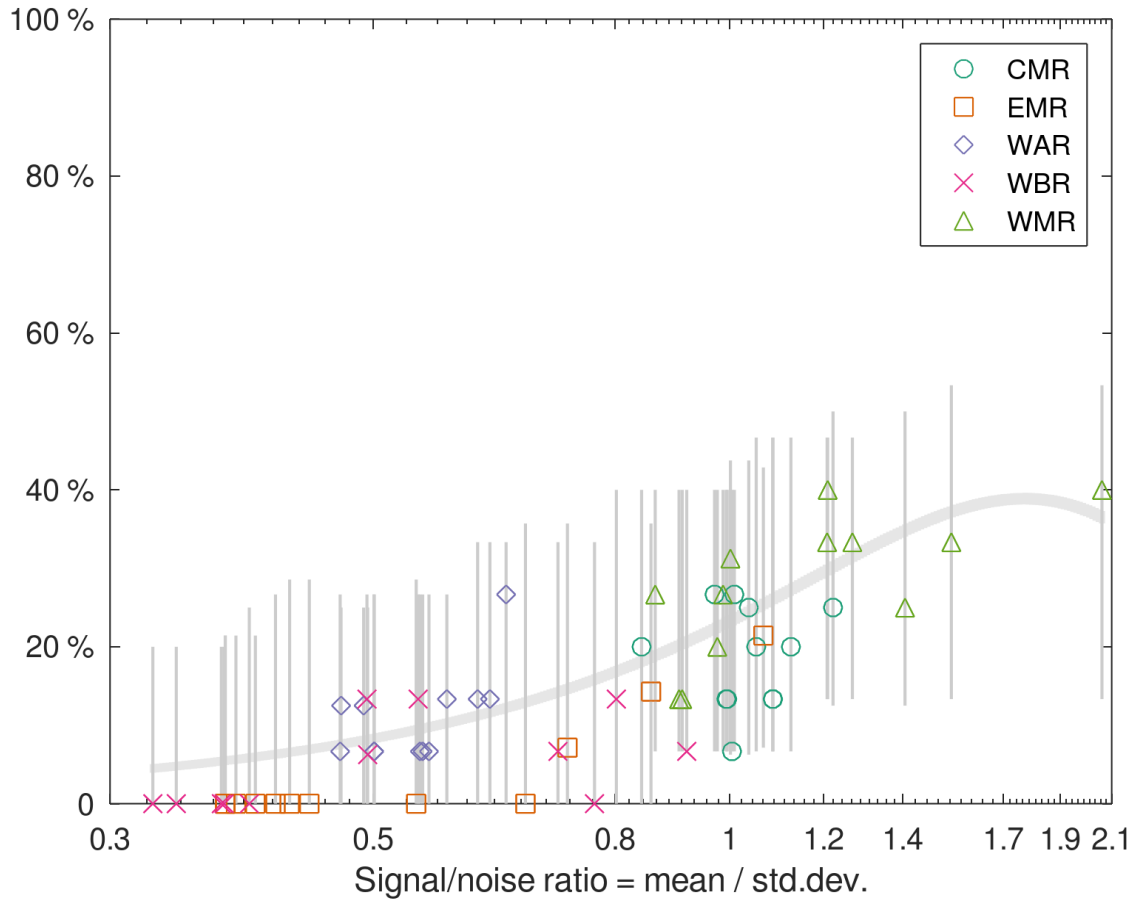

Figure S.19: frequency of values in the class "0.5" per each month and route (vertical axis) along with the corresponding signal-noise ratio (SNR, horizontal axis).The uncertainty of 10000 Monte Carlo simulations using a simplified approximation (Weibull distribution) for the unknown actual distribution is represented with the dark-grey vertical bars associated to each point (90 % of the total uncertainty, range between quantiles 5 % and 95 %, after rounding the simulated values to their nearest integer). The pale-grey vertical bars show a reference Weibull distribution (N=10000 instead of the much smaller sample size of each route in each month). Again, the 90 % uncertainty range is represented.

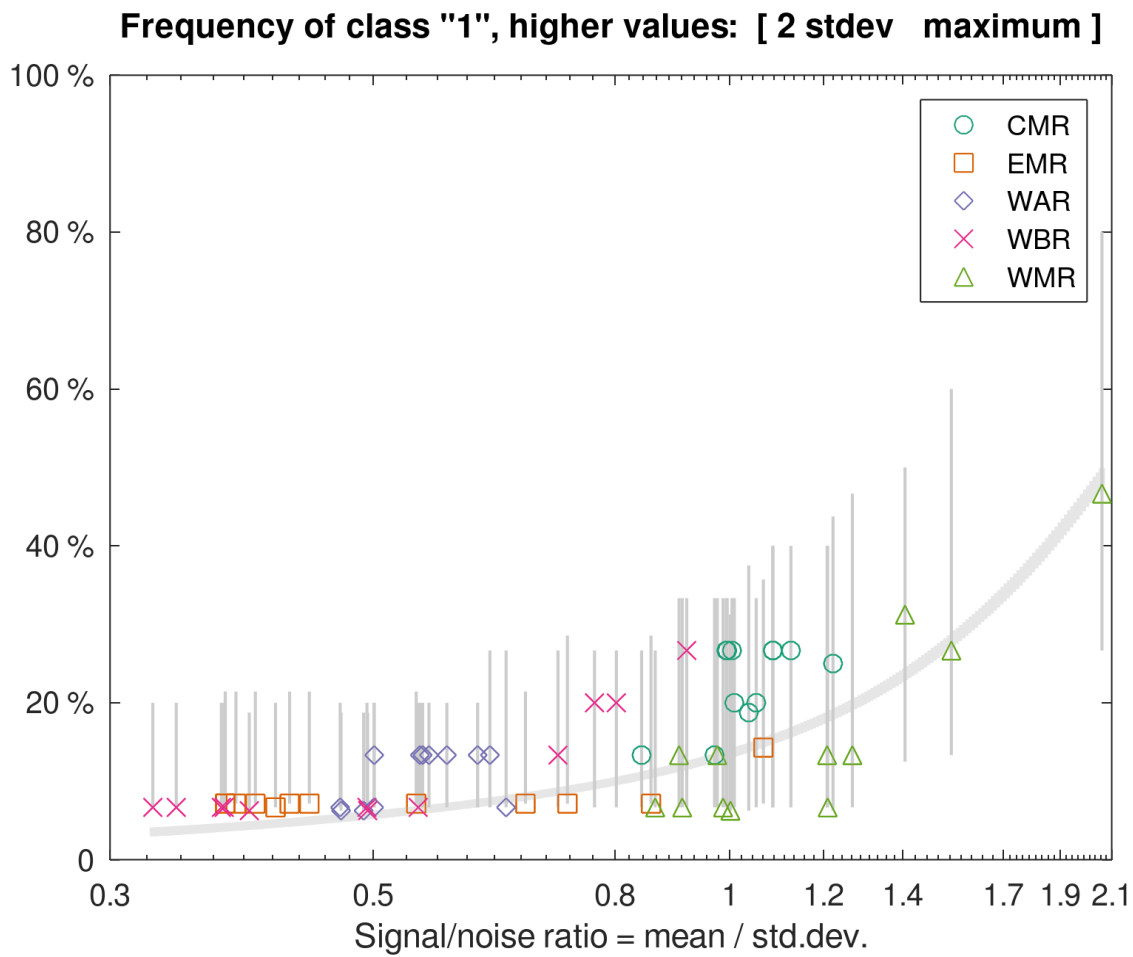

*Figure S.20: frequency of values in the class "1" per each month and route (vertical axis) along with the corresponding signal-noise ratio (SNR, horizontal axis).The uncertainty of 10000 Monte Carlo simulations using a simplified approximation (Weibull distribution) for the unknown actual distribution is represented with the dark-grey vertical bars associated to each point (90 % of the total uncertainty, range between quantiles 5 % and 95 %, after rounding the simulated values to their nearest integer). The pale-grey vertical bars show a reference Weibull distribution (N=10000 instead of the much smaller sample size of each route in each month). Again, the 90 % uncertainty range is represented.*

## S.7. Analytical framework for elaborating considerations on IBCs future trends

For the estimation of future IBCs, expert analysts provide an evidence-informed expectation on the future developments in the coming months, based on their assessment of the past and current situation (e.g. in relation to key trends and factors influencing IBCs), and the likelihood of future developments (see Figure S.21). Analytical considerations provided by analysts are thus used to assign a class to the future observations and generate the forecasts.

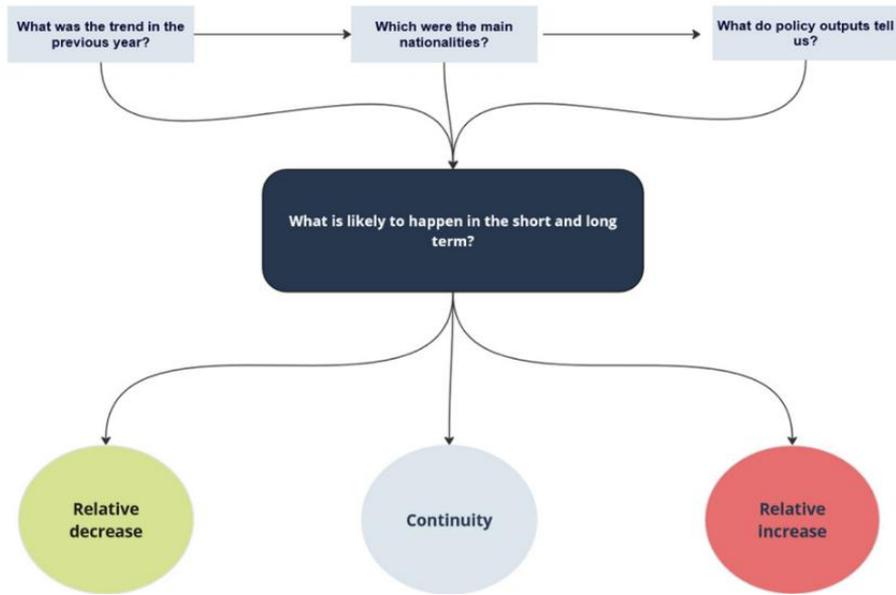

*Figure S.21: Analytical framework for elaborating considerations on IBCs future trends.*